\documentclass{article}

\usepackage{style/macros}
\usepackage[capitalize,noabbrev,nameinlink]{cleveref}

\author{Joseph Turnbull$^{1}$\
Jamie Simon$^{1,2}$   \\
\\[-0.5ex]
\texttt{\{joeyturnbull,jsi\}@berkeley.edu}\\
}

\newif\ifarxiv
\arxivtrue

\hypersetup{pdftitle={You Don't Need Feature Learning to Do In-Context Learning}}

\title{You Don't Need Feature Learning to Do In-Context Learning}

\usepackage{style/iclr2027_conference,times}
\iclrfinalcopy % Uncomment for camera-ready version, but NOT for submission.

\begin{document}
\maketitle

\begingroup
    \let\thefootnote\relax
    \footnotetext{\hspace{-6pt} $^1$UC Berkeley \ $^2$Imbue.}
    \footnotetext{\hspace{-2pt} Code: \url{https://github.com/JoeyTurn/kernels_can_icl}}
\endgroup
% \begingroup
%     \let\thefootnote\relax
%     \footnotetext{\hspace{-6pt} $^\circ$UC Berkeley \ $^\imbue$Imbue}
%     \footnotetext{\hspace{-2pt} Code: \url{https://github.com/JoeyTurn/kernel-icl}}
% \endgroup

\begin{abstract}

% I thought it was nice to frame this intro question in a broader way than you did below. I also think this is more the question you actually go after with this project: when you ask ``what separates a model that... from a model that...'' I come in expecting you to deliver something like post-hoc metrics that indicate which one the model learned. In fact, you're actually going after a broader conceptual question, and I find it curiosity-inspiring to ask that question directly.
% \remind \joey{Maybe change first sentence to introduce the paradox instead of just asking the question}

% In-context learning (ICL) presents a paradox: the training objective encourages memorizing a fixed set of tasks, and yet models still learn to generalize to new ones...

% memorizing tasks, yet models still learn to generalize to new ones.

How should we understand memorization vs. generalization in in-context learning (ICL) tasks, and when does one happen vs. the other?
To investigate, we study how rotation-invariant kernel machines learn the ICL linear regression task proposed by \citet{raventos2023pretrainingtaskdiversityemergence}.
By decomposing the task in the kernel eigenbasis, we obtain a spectral picture of learning in this ICL setting.
This picture reveals that as sample size grows, kernel machines learn successively-higher-degree components of the memorizing solution, which depend upon successively-higher-degree statistics of the training task set.
The generalizing solution is approximately a low-degree truncation of the memorizing solution, and so the model first generalizes, then degrades into memorization as training proceeds.

\vspace{2mm}

The takeaways are threefold.
First, if this case is indicative, learning a ``generalizing'' solution on an ICL task is really just underfitting the memorizing solution!
Second, the fact that kernel machines can generalize on this task means that feature learning is inessential to ICL.
Third, transformers empirically trained to do ICL are shown to memorize tasks through the higher-order statistics of the training tasks that the kernel-based theory predicts, suggesting it applies even to models capable of feature learning.

% \vspace{2mm}

% % Count em!
% The takeaways are twofold.
% % Roll title!
% First, kernel machines can generalize on ICL tasks, meaning that feature learning is inessential for ICL.
% % I personally think it's okay to be casual here -- i.e., ``really'' and the exclamation mark. The point of formality is that it helps with clarity, and IMO this is clear.
% % Thought: ``task'' may be preferable to ``case.''
% Second, if this case is indicative, learning a ``generalizing'' solution on an ICL task is really just underfitting to the memorizing solution!
% % NO CIRCUITS

\end{abstract}

\section{Introduction}
\label{sec:intro}

% \joey{LEFTOVER NOTES: 1. do we footnote diffusion models? we should
% 2. i need to update fig 1
% 3. figure out where RF models go, probably appendix
% 4. make task space stuff more clear
% 5. appendix sections for experimental details need writing
% 6. i want to show you can generalize at T=d+1 if you whiten
% }

% no math here at all, only words
% - one of the most interesting things LLMs can do is to "figure out what the task is" from context. for example, an LLM told to write a new file for a software repo can match style and content to existing files present in its context.
% - this is called few-shot learning or ICL in the literature.

\begin{figure}[t]
    \centering
    \resizebox{\textwidth}{!}{%
      \includegraphics[height=4cm]{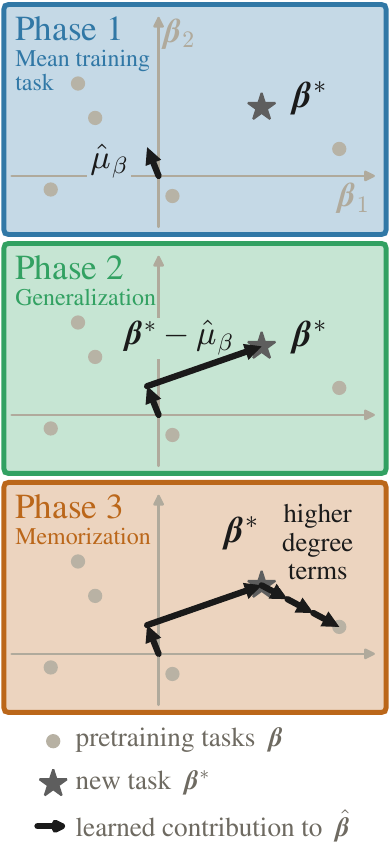}%
      \hspace{0.1cm}%
      \includegraphics[height=4cm]{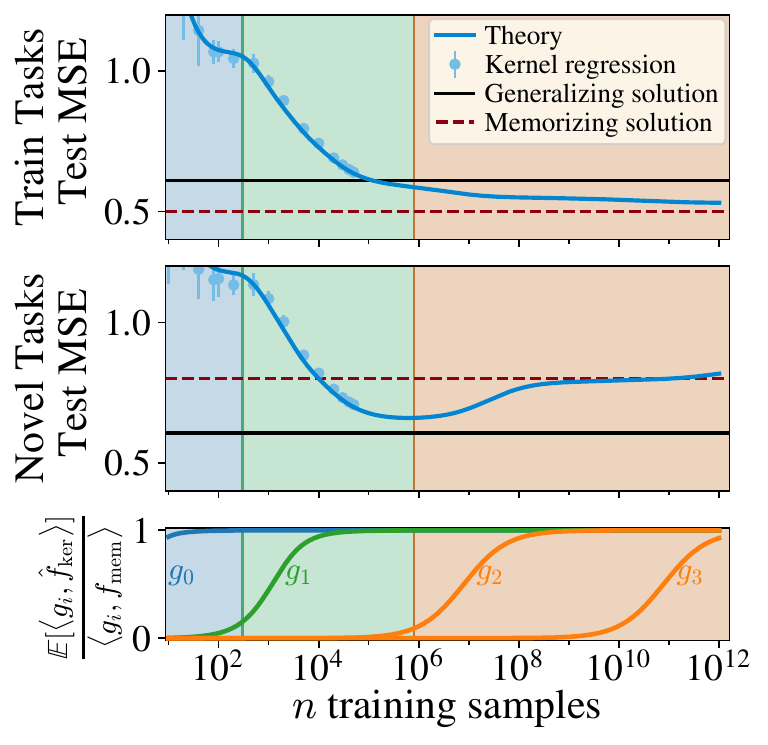}%
    }
    \definecolor{phaseoneblue}{HTML}{3058a6}
    \definecolor{phasetwogreen}{HTML}{218152}
    \definecolor{phasethreeorange}{HTML}{bb6719}
    \caption{
    \textbf{As sample size grows, kernels learn successively higher-degree target function components in in-context linear regression. They first learn the mean training task (\textcolor{phaseoneblue}{phase 1}), then approximate the degree-three generalizing solution (\textcolor{phasetwogreen}{phase 2}), and finally memorize the training tasks (\textcolor{phasethreeorange}{phase 3}).}
    \textbf{(Left)} A visualization of a kernel machine's inferred task in task space $\taskspace = \R^d$ decomposed into vector components learned in each phase (\textbf{arrows}).
    \textcolor{gray}{Grey dots} indicate training tasks $\taskspacetr = \{\vbeta^{(t)}\}$ and a \textcolor{black!80}{grey star} indicates an unseen task $\vbeta^* \notin \taskspacetr$.
    In \textcolor{phaseoneblue}{phase 1}, the kernel learns the mean training task $\hat\mu_\vbeta := \E{\vbeta \sim \taskspacetr}{\vbeta}$.
    In \textcolor{phasetwogreen}{phase 2}, the kernel learns approximately $\vbeta^*-\hat\mu_\vbeta$, nearly recovering the true task $\vbeta^*$.
    In \textcolor{phasethreeorange}{phase 3}, learned higher-degree terms pull the prediction towards the nearest task in $\taskspacetr$.
    \textbf{(Right)}
    We plot learning curves for KR on the in-context linear regression problem with $d=5,L=7,T=16$.
    \textbf{(Top)} Evaluated on the training tasks, the error falls monotonically, eventually reaching that of the memorizing solution (\cref{eq:memdef}).
    We show both theoretical predictions (lines) and experimental values (dots) in the experimentally-accessible region $n \in [1, 5 \times 10^4]$.
    \textbf{(Middle)} Evaluated on novel tasks ($\vbeta \sim \mathcal{N}(0, \mI_d)$), error is minimized at the end of \textcolor{phasetwogreen}{phase 2}, approaching that of the generalizing solution (\cref{eq:gendef}), before rising to that of the memorizing solution in \textcolor{phasethreeorange}{phase 3}.
    Background shading indicates the three phases of learning.
    \textbf{(Bottom)}
    We also plot the predicted overlap between the function $\hat{f}_{\mathrm{ker}}$ learned by KR and various polynomial functions $g_i$.
    The functions $g_i$ correspond to the vector components in the cartoon on the left, and are given explicitly in \cref{app:experiment_details}.
    The function $g_i$ is determined by the $i$-th order statistics of $\taskspacetr$.
    The component $g_1$ is responsible for generalization and is learned in \textcolor{phasetwogreen}{phase 2}.
    % We additionally predict how many samples are needed for a kernel to learn each relevant function $\phi_i$ (with multiple high order terms shown for \textcolor{phasethreeorange}{phase 3}).
    % See \cref{app:experiment_details} for experimental details.
    %
    We show theoretical predictions (lines) for the kernel's error out to $10^{12}$ samples, finding excellent agreement between theory and experiment in the experimentally accessible region.
    }
    \label{fig:summary}
\end{figure}

Large language models (LLMs) exhibit the striking ability to infer an intended task from context.
% don't nec need this sentence
For example, an LLM told to write a new file in a software repository can match style and content to existing files present in its context.
More generally, an LLM given a few examples of a desired task can often infer the task and apply it to new settings.
% \joey{Both few- and zero-shot as I've come to learn}
This is called few-shot learning or \textit{in-context learning} (ICL) in the literature.
This useful behavior and the conditions under which it arises remain largely mysterious.\footnote{We want to mention that ICL is often, but not always, wanted. For example, an ideal model given false/adverse information would rely on memorized facts instead of providing a context-driven output.}

How should we begin to describe ICL in mathematical terms?
A general setting is as follows.
Suppose that there exists a set $\taskspace$ of all possible tasks, of which only a subset $\{\vbeta^{(t)}\}_{t=1}^T =: \taskspacetr \subset \taskspace$ are present in the model's training data.
We might suppose that each task specifies a certain function $f_\vbeta: \X \rightarrow \Y$ between two spaces.
A given \textit{context} $\vz \in (\X \times \Y)^L \times \X$ consists of a number of examples $(x_i, y_i)_{i=1}^L$, where $y_i = f_\vbeta(x_i) + \eta_i$ with noise $\eta_i$, together with an unpaired input $x_{L+1}$.
The model is trained to map $\vz \mapsto f_\vbeta(x_{L+1})$, which requires first inferring $\vbeta$ from the example pairs and then applying it to the unpaired input $x_{L+1}$.
The most interesting questions arise when the model is provided many examples for each of the $T$ tasks at training time, but at test time is provided a \textit{new} task $\vbeta^* \notin \taskspacetr$.
A model performing ICL will be able to extract the task and apply $f_{\vbeta^*}$ to the unpaired input.
We would like to know when the model can perform well on unseen tasks --- and when it can, to understand why, and how it learned to do so.
% \jamie{changed $\vx_i$ to $x_i$ here bc we're still in a general space} %sg

It is worth stressing that generalization in ICL is strange indeed, and is in fact rather paradoxical.
The Bayes-optimal solution for the training objective is to always interpret any new context $\vz$ as indicating one of the $T$ training tasks and to simply apply that task.
(Or, if there are several plausible candidates, as a distribution over several training tasks.)
Given this training objective, it would seem hopeless that the model would correctly infer the unseen task $\vbeta^* \notin \taskspace_\mathrm{tr}$.
Nevertheless, it evidently does so in many practical and useful settings.
Our work in this paper is to offer a resolution to this paradox and to explain how, in spite of this mismatch, generalization occurs in ICL.

The main idea is that \textbf{generalization occurs mid-optimization, as a milestone on the road to memorization.}
It is well-appreciated that neural networks learn simple functions early in training and more complex functions later on (e.g. \citet{rahaman:2019-spectral-bias,nakkiran2019sgdneuralnetworkslearns,xu:2019-frequency-principle-A}).
We suggest that the generalizing solution is in fact just a simpler version of the memorizing solution, stopped before it learns the higher-degree functional components required for exact memorization of the training set.
We arrive at these conclusions from a study of the ICL-for-linear-regression setting of \citet{raventos2023pretrainingtaskdiversityemergence}.
In this setting, $\X = \R^d$, $\Y = \R$, $\taskspace = \R^d$, and each task is simply a linear map $f_\vbeta(\vx) = \vbeta^\top \vx$.
Task vectors are sampled $\vbeta \sim \N(0,\mI_d)$.
The beauty of this setting is that the Bayes-optimal train-time (memorizing) and test-time (generalizing) solutions can be written down in closed form (\cref{eq:gendef,eq:memdef}).
% This simplicity yields exact analytical expressions for the memorizing and generalizing solutions, either of which a model could hypothetically use.
% 
Which solution a model learns is largely determined by the model's inductive biases.
This suggests studying \textit{kernel regression} (KR) as the model, since there exists a well-developed spectral theory of its inductive bias \citep{bordelon:2020-learning-curves,jacot:2020-KARE,simon:2021-eigenlearning}.
The main technical hurdle is in computing the kernel eigendecomposition on this task so that this existing theory may be applied.

Surmounting this hurdle, we find a resolution to the aforementioned paradox:
the generalizing solution is approximately a \textit{low-polynomial-degree truncation of the memorizing solution.}
%
% Kernel machines gradually learn the memorizing solution one term at a time, arriving at the \textit{generalizing solution} at an intermediate point in training.
As the dataset size grows, kernel machines first learn simple low-degree polynomial components of the memorizing function before resolving increasingly high-degree components.
They approximate the generalizing solution after learning the degree-three components of the memorizing solution.
Given enough samples, they will proceed to fully memorize.
Our results put theory behind the empirics of \citet{raventos2023pretrainingtaskdiversityemergence}, giving the first end-to-end theory of the generalization-memorization transition in ICL.
This theory explains the ``transience'' phenomenon in which a model first learns the generalizing solution before ``forgetting it'' in favor of the memorizing solution \citep{singh2023transientnatureemergentincontext}. 

% % 
% Despite this fact, prior work \citep{singh2023transientnatureemergentincontext} has shown that transformers seemingly implement a solution that's Bayes-optimal on the prior distribution from which the training tasks were drawn, labeled as ``transience of ICL''.
% % 
% This same phenomenon was observed in MLPs \citep{tong2025mlpslearnincontextregression}, suggesting that the model architecture is not the cause.
% % 
% What, then, can be causing this behavior?
% % 
% This explains the \textit{ICL transience} phenomenon in which a model first learns the generalizing solution before apparently ``forgetting it'' in favor of the memorizing solution \citep{singh2023transientnatureemergentincontext}.

% Each step of the learning process can be cast in terms of statistics of $\taskspacetr$, with earlier steps dependent only on the lower-order moments.
% %
% This explains the \textit{ICL transience} phenomenon in which a model first learns the generalizing solution before apparently ``forgetting it'' in favor of the memorizing solution \citep{singh2023transientnatureemergentincontext}.
% \jamie{I actually don't see how the first sentence of this para connects to / implies the second one at all. can we just cut this para entirely? alternatively, cut the first sentence and add the second sentence}

A surprising consequence of our study of fixed-feature kernel machines is that \textbf{a model does not need feature learning to do ICL}.
(Of course, feature learning certainly allows a model to do ICL more \textit{efficiently.})
This suggests that memorization and generalization in ICL can be understood in simple polynomial language, without reference to feature learning or circuit formation.
Experiments on feature-learning transformer models (\cref{sec:transformers}) show that our kernel-based theory can be used to substantially delay when the models memorize, by removing higher-order statistics of the training tasks at fixed task diversity.
This refines the task-diversity picture of \citet{raventos2023pretrainingtaskdiversityemergence}: what matters for generalization is how Gaussian the training tasks are, not how many there are.

\section{Research context and related works}
\label{sec:related_works}

Our work sits at the intersection of four threads: the tension between memorizing and generalizing Bayes-optimal solutions (most developed for diffusion models), in-context learning dynamics given finite task diversity, simplicity biases in learning, and the eigenstructure theory of kernel regression.

\paragraph{Memorizing and generalizing Bayes-optimal solutions.}
Whenever a model is trained on a finite sample from some latent distribution, two Bayes-optimal predictors are available: one that treats the empirical distribution as the prior (memorizing), and one that uses the population prior (generalizing).
The former generically takes the form of a softmax-weighted average over training points.
In our setting, the memorizing solution is the dMMSE (\cref{eq:memdef}), a posterior average over the $T$ pretraining tasks in \textit{task space}.
In diffusion modeling, the memorizing solution is akin to a nearest-neighbor match to the closest training image \citep{scarvelis2025closedformdiffusionmodels,Biroli_2024,kamb2025analytictheorycreativityconvolutional}.
In dense associative memories, retrieval is likewise a softmax-weighted average over stored patterns \citep{krotov2016denseassociativememorypattern,ramsauer2021hopfieldnetworksneed}.
The same paradox arises in each case: %
fully optimizing the training objective gives a predictor that cannot produce anything novel, so any observed generalization must come from the learner \emph{failing} to reach that optimum.
The closest analogue to our resolution of this paradox comes from the diffusion literature.
\citet{li2024understandinggeneralizabilitydiffusionmodels} found that linear models capturing the same Gaussian structure as nonlinear diffusion models are nearly identical to the optimal denoiser for Gaussian data.
\citet{bonnaire2025diffusionmodelsdontmemorize} train a random Fourier features model to show that memorization can be avoided through early stopping, given a strong enough implicit bias.
\citet{kamb2025analytictheorycreativityconvolutional,lukoianov2025localityimagediffusionmodels} similarly attribute the avoidance of memorization in diffusion models to implicit biases of the network (equivariance) and the data (locality).
Together, these results parallel our finding that the generalizing solution is a truncation of the memorizing one at its Gaussian statistics (\cref{prop:cumulant-truncation}), implemented through the inductive biases of the learner.

\paragraph{In-context learning.}
Since its description by \citet{brown-2020-gpt3}, ICL has remained a somewhat enigmatic capability of language models.
To understand it, researchers have studied model-side attributes, such as parameter count, depth and the resulting expressivity of the architecture \citep{brown-2020-gpt3,wei2022emergentabilitieslargelanguage,garg2023transformerslearnincontextcase}, as well as data-side attributes, such as task diversity \citep{chan2022datadistributionalpropertiesdrive,kirsch2024generalpurposeincontextlearningmetalearning,raventos2023pretrainingtaskdiversityemergence,park2025competitiondynamicsshapealgorithmic}.
Closest to our non-attention setting, \citet{tong2025mlpslearnincontextregression} show that MLPs perform ICL linear regression in a similar manner to transformers.
On the theoretical side, most analyses of ICL have focused on linear attention \citep{Lu_2025,letey2025pretraintesttaskalignmentgoverns,Cui_2024,li2025gatingweightingunderstandinggated}.
These models are tractable and give an end-to-end description of learning, but they cannot express the memorizing solution, and so cannot capture the transition toward it: they are, by construction, unable to exhibit transience.

\paragraph{Transience.}
\citet{singh2023transientnatureemergentincontext} found that ICL can emerge and later vanish during training.
Most explanations of this transience posit competing memorizing and generalizing circuits \citep{carroll2025dynamicstransientstructureincontext,singh2025strategycoopetitionexplainsemergence,nguyen2025differentiallearningicl}.
Others \citep{wurgaft2025incontextlearningstrategiesemerge,park2025competitiondynamicsshapealgorithmic} claim that transience can be explained through a linear combination of the memorizing and generalizing solutions.
In \cref{subsec:transience}, we explain transience from \textit{first principles} using \textit{fixed-feature} models.
To be clear, we give the first description of transience purely as a consequence of our theory---we make no curve fits on training trajectories, and we find transience even without models having the ability to learn circuits.

\paragraph{Kernel ridge regression as a model of learning dynamics.}
Kernel methods have proven useful as models of network dynamics and optimization \citep{chizat:2019-lazy-training,du:2019-convergence,bordelon:2021-sgd}, thanks in large part to the neural tangent kernel (NTK) line of work \citep{neal:1996,lee:2018-nngp,jacot:2018}, in which suitably initialized wide networks train like kernel learners.
Recent works show that the test error of kernel learners can be predicted from the \textit{kernel and task eigenstructure} \citep{sollich:2001-mismatched,bordelon:2020-learning-curves,jacot:2020-KARE,simon:2021-eigenlearning}.
This ``KRR eigenframework'' depends on the kernel's eigenvalues and eigenfunctions with respect to the data distribution, and is reviewed in \cref{app:krr_eigenframework}.

\paragraph{Simplicity biases and statistics of increasing order.}
Neural networks are known to fit low-frequency or low-complexity functions first \citep{nakkiran2019sgdneuralnetworkslearns,abbe2024mergedstaircasepropertynecessarynearly}.
This suggests that the low-to-high ordering of learning task-space statistics we find for kernels may also hold for feature-learning networks.

\paragraph{Approximating the kernel eigenstructure.}
Our eigensystem-finding framework (\cref{def:see}) follows the strategy of \citet{karkada2026predictingkernelregressionlearning}, by approximating the kernel as an inner-product kernel, creating subspaces of polynomials, and performing Gram-Schmidt against the data distribution.
Our work trades understandability for expressivity: we cannot generally get closed form expressions for the eigenfunctions, but our framework is more widely applicable.

% \paragraph{Exactly solved cases of kernel eigenstructure.}
% Exact kernel diagonalizations for machine-learning-relevant kernels are known in several highly symmetric settings.
% %
% These include stationary kernels on the torus $\mathbb{T}^d$ and rotation-invariant kernels on the sphere $\mathbb{S}^d$ \citep{mei:2019-double-descent}, rotation-invariant kernels on a product-of-spheres measure \citep{ghorbani:2020-when-nets-outperform-kernels}, and rotation-invariant kernels on anisotropic Gaussian data \citep{karkada2026predictingkernelregressionlearning,wortsman2025kernelridgeregressionpowerlaw}.
% %
% Our subkernel expansion framework (\cref{def:see}) follows the strategy of \citet{karkada2026predictingkernelregressionlearning}, building eigenfunctions level by level from low-order data statistics.
% %
% The construction can be applied to any data measure, though we validate it only on ICL data, and it is controlled when the level coefficients separate (\cref{appsubsec:sex_proof}).
% %
% For a single task ($T=1$), ICL data is Gaussian and the subkernel expansion framework reduces to their Hermite structure (\cref{appsubsec:sexpansion}).
\section{Preliminaries}
\label{sec:prelim}

% \jamie{the Preliminaries section is for introducing concepts and notation that you will use in the paper.}

\textbf{In-context linear regression.}
We study the in-context linear regression setting of \citet{raventos2023pretrainingtaskdiversityemergence}, which is defined as follows.
At the outset, we construct a pool of task vectors $\taskspacetr := \{\vbeta^{(t)}\}_{t=1}^T$ of $T$ tasks sampled $\vbeta^{(t)} \sim \N(0, \mI_d)$ to generate the training data.
In this setting, one training sample is a \textit{context vector} constructed as follows.
First, a task $\vbeta$ is drawn uniformly from $\taskspacetr$.
Second, $L+1$ vectors $(\vx_i)_{i=1}^{L+1}$ are sampled $\vx_i \sim \N(0, \frac{1}{d} \mI_d)$.%
\footnote{Anisotropic distributions for both $\vx$ and $\vbeta^{(t)}$ are discussed in \cref{app:correspondence,app:eigensystem,app:oracles}.}
Third, $L+1$ values $(y_i)_{i=1}^{L+1}$ are computed as $y_i = \vbeta^\top \vx_i + \eta_i$, with noise $\eta_i \sim \N(0, \epsilon^2)$.
Finally, we construct the context
\begin{equation}
    \vz = \left[{\vx_1}^\top, y_1, \ldots, {\vx_L}^\top, y_L, {\vx_{L+1}}^\top\right] \in \mathbb{R}^{(d+1)L+d},
\end{equation}
with $y_{L+1} =: y$ held out as the regression target.
We will write $\d_\vbeta$ for the joint distribution of $(\vz, y)$ conditioned on a given task vector $\vbeta$.
We refer to $T$ as the \textit{task diversity} and $L$ as the \textit{context length}.

\textbf{Kernel regression.}
We work in a standard supervised setting in which the model is trained on $n$ contexts
$\mathcal{Z} = \left\{\vz^{(i)}\right\}_{i=1}^n$ and the corresponding scalar targets $\vy = \left( y^{(i)} \right)_{i=1}^n$.
We take KR as the learning algorithm.
KR is specified by a choice of positive-semidefinite kernel function $K(\vz,\vz')$ and a choice of ridge parameter $\delta \ge 0$.
We consider various rotation-invariant kernel functions: $K(\vz, \vz') = K(\vz^\top \vz', \norm{\vz}, \norm{\vz'})$.
KR returns the predicted function $\hat{f}(\vz) = K_{\vz \mathcal{Z}} \left( K_{\mathcal{Z} \mathcal{Z}} + \delta \mI_n \right)^{-1} \vy$, where $[K_{\vz \mathcal{Z}}]_{m} = K(\vz, \vz^{(m)})$ and $[K_{\mathcal{Z} \mathcal{Z}}]_{km} = K(\vz^{(k)}, \vz^{(m)})$.

\textbf{Evaluating the learned function.}
Once we have the learned function $\hat{f}$, this will be evaluated two ways, on two distributions.
First, we study its error on new contexts from the \textit{training task distribution:} $\d_\mathrm{tr} := \frac{1}{T} \sum_{\vbeta \in \taskspacetr} \d_{\vbeta}$.
Second, we compute its error on the \textit{prior distribution}, with contexts generated from fresh tasks: $\d_\mathrm{novel} := \E{\vbeta \sim \N(0, \mI_d)}{\d_{\vbeta}}$.
We compute mean-squared error (MSE) over both distributions:
\begin{equation}
    \e_{\mathrm{tr}}(\hat{f}) := \E{(\vz, y) \sim \d_{\mathrm{tr}}}{\left(\hat{f}(\vz) - y\right)^2},
    \qquad
    \qquad
    \e_{\mathrm{novel}}(\hat{f}) := \E{(\vz, y) \sim \d_{\mathrm{novel}}}{\left(\hat{f}(\vz) - y\right)^2}.
\end{equation}

\textbf{Memorizing and generalizing solutions}
The \textit{memorizing solution} is defined as that which minimizes error on the training distribution: $f_\mathrm{mem} := \mathrm{argmin}_f \e_\mathrm{tr}(f)$.
As shown by \citet{raventos2023pretrainingtaskdiversityemergence}, $f_\mathrm{mem}$ takes the discrete-minimum-mean-squared-error (dMMSE) form
\begin{equation}
    f_\mathrm{mem}(\vz) = \frac{
    \sum_{t=1}^T w_t(\vz) \cdot \vbtt\vx_{L+1}
    }
    {
    \sum_{t=1}^T w_t(\vz)
    },
    \;\;\;
    \text{where}
    \;\;\;
    w_t = \exp\left(-\frac{1}{2\epsilon^2}\sum_{i=1}^{L}\left(y_i-\vbtt\vx_i\right)^2\right).
    \label{eq:memdef}
\end{equation}
The \textit{generalizing solution} is the Bayes-optimal solution on the prior distribution: $f_\mathrm{gen} := \mathrm{argmin}_f \e_\mathrm{\mathrm{novel}}(f)$.
This is simply ridge regression on the context:
\begin{equation}
    f_\mathrm{gen}(\vz) = \vx_{L+1}^\top \left( \mX \mX^\top + \epsilon^2 \mI_d \right)^{-1}\mX \vy_{1:L},
    \qquad
    \mathrm{where}
    \ \ 
    \mX = [\vx_1, \cdots, \vx_L].
    \label{eq:gendef}
\end{equation}

We will find that, as sample size $n$ grows, KR first learns $f_\mathrm{gen}$, then learns $f_\mathrm{mem}$ as $n \rightarrow \infty$.

\textbf{Moments and cumulants.}
We write $\hat{\vmu}_\vbeta := \frac{1}{T} \sum_t \beta^{(t)}$ for the mean of the training tasks $\taskspacetr$ and $\hat\Sigma_\vbeta$ for its covariance.
We write $\hat\kappa_\ell$ for the order-$\ell$ cumulant tensor of $\taskspacetr$.
\section{The kernel eigenstructure}
\label{sec:eigenstructure}

\subsection{Review of the KR eigenframework}
Prior work has obtained a predictive analytic framework which gives the generalization behavior of KR on essentially arbitrary distributions (see e.g. \citet{simon:2021-eigenlearning}).
This \textit{KR eigenframework} takes as inputs the kernel eigensystem and the associated eigendecomposition of the target function.
The eigensystem of a kernel function $K(\cdot, \cdot)$ w.r.t. a distribution $\d$ is the set of scalar-function pairs $\lambda_i, \phi_i$ that satisfy $\E{\vz' \sim \d}{K(\vz, \vz') \phi_i} = \lambda_i \phi_i$.
A function $f^*$ is then decomposed into coefficients $v_i = \mathbb{E}_{z\sim \d} [\phi_i(z)f^*(z)]$.
The KR eigenframework gives equations predicting quantities such as $\E{}{\hat{f}} \approx \sum_i v_i\phi_i$ and the test-time MSE.
The main qualitative idea is that the target components $v_i \phi_i$ are learned in decreasing order of $\lambda_i$: eigenfunctions with larger eigenvalue are learned first.
This behavior is known as \textit{spectral bias.}

\subsection{The kernel eigensystem of in-context linear regression}
In order to use the KR eigenframework in our ICL setting, we need to compute the task eigensystem and target eigencoefficients (or at least say enough to draw useful conclusions).
The training data are sampled $\vz \sim \d_\mathrm{tr}$ and have target function $\E{}{y | \vz} = f_\mathrm{mem}(\vz)$.
We thus need to eigendecompose a rotation-invariant kernel $K$ w.r.t. distribution $\d_\mathrm{tr}$.

This eigensystem cannot generally be found in closed form, so to make progress, we invoke a series of assumptions and approximations.
We give a sketch here and a derivation in \cref{app:eigensystem}.
First, following \citet{karkada2026predictingkernelregressionlearning}, we approximate the kernel as an \textit{inner-product kernel} $\tilde{K}(\vz, \vz') = \sum_{\ell \ge 0} \frac{c_\ell}{\ell!} (\vz^\top \vz')^\ell$, obtaining the constants $\{c_\ell\}$ by matching $K$ and $\tilde{K}$ on the sphere $r \cdot \mathbb{S}^{\mathrm{dim}(\vz)-1}$ with $r = \mathrm{rms}\left[\norm{\vz}\right]$.
Then, as in that same paper, we assume $c_\ell \gg c_{\ell+1}$, which implies that the kernel eigensystem can be found by:
\begin{enumerate}
    \item taking the subspaces $(\mathcal{P}_{\le \ell})_{\ell=0}^\infty$, where $\mathcal{P}_{\le \ell}$ contains all polynomials of $\vz$ up to order $\ell$;
    \item performing Gram-Schmidt on these subspaces in order of increasing $\ell$ to obtain orthogonal subspaces $(\tilde{\mathcal{P}}_{\le \ell})_{\ell=0}^\infty$; and finally
    \item diagonalizing each subspace $\tilde{\mathcal{P}}_{\le \ell}$ w.r.t. $\d_\mathrm{tr}$.
\end{enumerate}
This procedure may appear more difficult and obscure than simply diagonalizing the empirical kernel numerically: %
for example, the $\ell$-th subspace has dimension $\left((d+1)L+d\right)^\ell$, which quickly exceeds the dimension $n$ of the empirical kernel's eigenspace for even moderate values of $d, L,$ and $\ell$. 
However, symmetries in the problem setup substantially reduce the complexity in every subspace.
Specifically, because the in-context sample-label pairs are permutation invariant, the subspaces can be Kronecker-factorized (reducing the effective dimension to roughly $(d+1)^\ell$) and further decomposed into an $L$-scaling rank-$d$ spike plus a bulk term.
% (particularly those that do not scale with $L$)
At large context lengths, the leading eigenfunctions can be found in closed form, while the remainder must be computed numerically (due to the unsolvability of the quintic).
Luckily, the most important sets of the eigenfunctions (those with non-zero target coefficients) are those that can be found in closed-form.
Despite being analytical, these functions are highly complex (see \cref{subsec:nogen,app:correspondence}).
Here, we give said eigenfunctions in a more comprehensible, reduced form:
% The most important (sets of) eigenfunctions that emerge anayltically are as follows:
% The two most important (sets of) eigenfunctions that emerge analytically in this way are as follows:
\begin{itemize}
    \item at order one, there are $d$ ``linear readout'' modes $\vphi_{1}(\vz) = \vx_{L+1}$, which have eigenvalue $\lambda_1 = \frac{c_1}{d}$;
    \item at order three, there are again $d$ eigenmodes (see \cref{apptable:eigcoeffs}), with the most significant for the kernel learner being a single ``regression'' component
    \begin{equation} \label{eq:phi_3}
        \phi_3(\vz) \approx \frac{d}{L+d+1}\left( \sum_{i=1}^L y_i \vx_i^\top - \tfrac{L}{d}\hat\vmu_\vbeta^\top \right) \vx_{L+1},
    \end{equation}
    which has eigenvalue $\lambda_3 \propto L\frac{c_3}{d^3}$ ($c_3/c_1\sim L^{-1}$). Note that $\sum_i y_i\vx_i^\top \approx \tfrac{L}{d} \vbeta^\top$ as $\sum_i\vx_i \vx_i^\top \approx \tfrac{L}{d}\mI_d$ for $L \gg d$;
    \item at order $2\ell+1$ ($\ell \ge 2$), there are $d^{\ell+1}/\ell!$ ``memorizing'' components taking the form
    \begin{equation} \label{eq:high_phi}
        \vphi_{2\ell+1}(\vz) \sim \frac{1}{L^{1+\ell}}\left(\sum_{i}^L y_i\vx_i -\dots\right)^{\otimes \ell} \vx_{L+1},
    \end{equation}
    where we have dropped many highly involved prefactors and denoted orthogonalization against lower orders with dots.%\footnote{The $\otimes_\ell$ in the }
\end{itemize}
In general, the top eigenvalues at order $\ell$ scale like $c_\ell \cdot L^{(\ell-1)/2}\cdot d^{-\ell}$ ($c_\ell$ scaling as $L^{-\ell/2}$).
With $c_\ell \gg c_{\ell+1}$, the target-relevant eigenfunctions are learned in ascending order.
We experimentally check our eigensystem against a true numerical eigensystem computed from a finite trainset for several common rotation-invariant kernels in \cref{app:eigensystem,fig:eigendecom}.
We see very good numerical agreement, which validates the assumptions we have made up to this point.
In particular, it confirms that these kernels will indeed learn target function components in order of increasing degree.

\subsection{Eigendecomposing the generalizing and memorizing solutions}

We must now find the eigencoefficients of $f_\mathrm{gen}$ and $f_\mathrm{mem}$.
We again sketch our results here and defer a full treatment to \cref{app:correspondence}.
We begin with decomposing $f_\mathrm{mem}$.
While the function appears highly complicated to extract coefficients from, we can use the law of total expectation (see \cref{lem:kernel-memorize}) to find the eigencoefficients without ever using $f_\mathrm{mem}$:
\begin{equation}
  v_{i,\mathrm{mem}} \equiv \mathbb{E}_{(\vz, y)\sim \mathcal{D}_\mathrm{tr}}[f_{\text{mem}}(\vz)\phi_i(\vz)]= \mathbb{E}_{(\vz, y)\sim \mathcal{D}_\mathrm{tr}}[y\phi_i(\vz)].
\end{equation}
Substituting $f_\mathrm{mem}$ for $y = y_{L+1}$ makes finding the eigencoefficients relatively simple.
Using $\mathbb{E}_{\vx \sim \mathcal{N}(0, \mI_d/d)}[y_k\vx_k^\top] \sim \mathbb{E}_{\vx \sim \mathcal{N}(0, \mI_d/d)}[\vbeta^\top \vx_k\vx_k^\top] \sim \vbeta^\top\mI_d/d$ for any $k$, we expect the order $2\ell+1$ eigenfunctions described earlier to carry eigencoefficients that are (to dropped prefactors) $\propto\mathbb{E}_{\d_\mathrm{tr}}[\vbeta^{\otimes \ell+1}]$.
In other words, the $2\ell+1$-th eigencoefficients are defined using up to the $\ell+1$-th cumulants of $\d_\mathrm{tr}$; cumulants as opposed to moments due to orthogonalization against lower order terms.
By symmetry, all even-order ($2\ell$-th) eigencoefficients of $f_\mathrm{mem}$ are zero.

$f_\mathrm{gen}$ can now be easily obtained.
$f_\mathrm{gen}$ can be decomposed in exactly the same way as $f_\mathrm{mem}$, with the distribution $\d_\mathrm{tr}$ changed to $\d_\mathrm{novel}$ ($\mathcal{N}(0, \mI_d)$).
The Gaussian distribution is the unique distribution with only a first and second cumulant.
Therefore, $f_\mathrm{gen}$ is understood as being primarily composed of the degree $3$ ``regression'' eigenfunction.
The degree $1$ ``linear readout'' is not included as we draw $\vbeta$ from a mean-$\boldsymbol{0}$ distribution.
Equivalently, one can observe that as $L/d \rightarrow \infty$,  $\mX \mX^\top \rightarrow \mathrm{const} \cdot \mI_d$ such that $f_\mathrm{gen}$ approaches the third-order polynomial $\vx_{L+1}^\top \mX \vy_{1:L}$.
$f_\mathrm{mem}$ truncated at order 3 approximates the exact same third-degree polynomial as $f_\mathrm{gen}$.
Thus, $f_\mathrm{gen}$ is approximately a degree-3 polynomial truncation of $f_\mathrm{mem}$.
Past the degree-3 truncation, $f_\mathrm{mem}$ contains higher-degree polynomials that do not exist in $f_\mathrm{gen}$, driven by the higher order statistics of the random, finite draws of $\vbeta^{(t)}$.

\subsection{The resulting picture of memorization and generalization}

\begin{figure}[t!]
    \includegraphics[width=\textwidth]{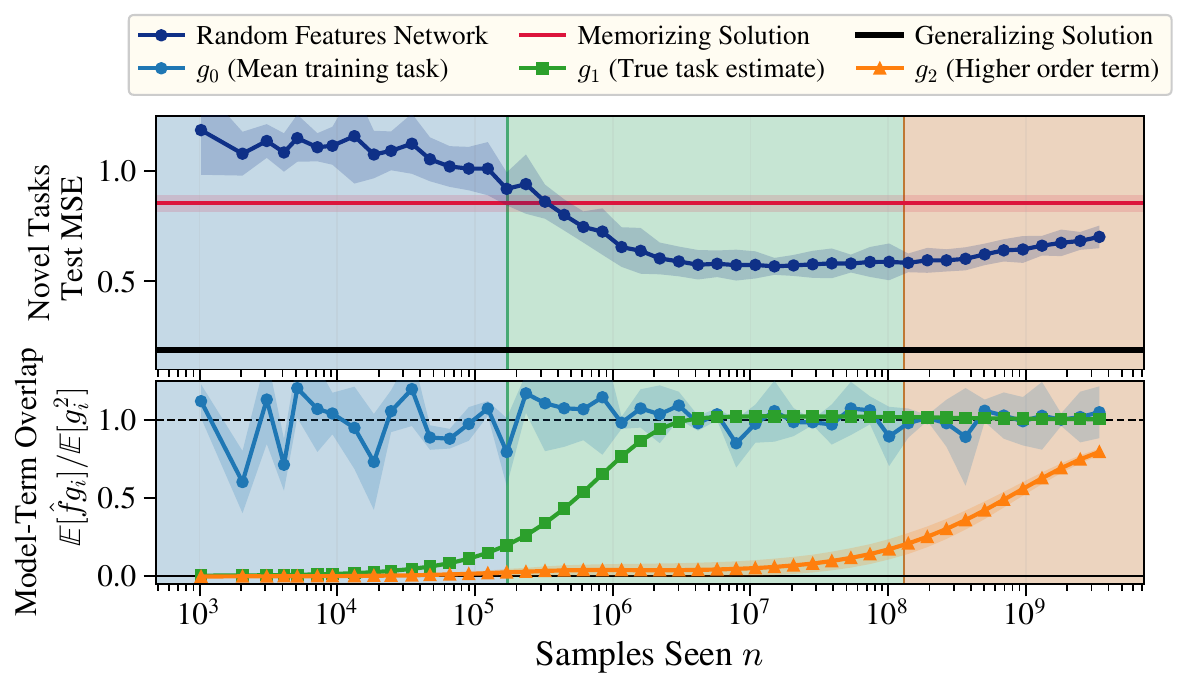}
    \definecolor{phaseoneblue}{HTML}{3058a6}
    \definecolor{phasetwogreen}{HTML}{218152}
    \definecolor{phasethreeorange}{HTML}{bb6719}
    \caption{
    \textbf{Transience is explained by learning higher order task statistics.}
    We train \textit{lazy} random features networks (RFNs) (see \cref{appsubsec:rf_analytical} for a description of RFNs) with online SGD on ICL linear regression data.
    Using RFNs, we empirically find that the error on novel tasks rises when the higher degree terms are learned, as was theoretically predicted in \cref{fig:summary}.
    We observe the same three phases of \textcolor{phaseoneblue}{predicting from the mean task alone}, then \textcolor{phasetwogreen}{generalizing to any task}, followed by \textcolor{phasethreeorange}{memorizing using higher-degree terms}.
    \textbf{(Top)} Test MSE on novel tasks $\beta \sim \d_\mathrm{novel}$, with the memorizing (\textcolor{red}{dMMSE}) and generalizing (\textbf{ridge}) oracles for reference.
    Unlike the kernels trained in \cref{fig:summary}, the random feature networks train with enough samples to empirically probe \textcolor{phasethreeorange}{phase 3}.
    \textbf{(Bottom)} Overlap of the learned predictor with the leading terms of the memorizing solution, $\mathbb{E}[\hat f g_i]/\mathbb{E}[g_i^2]$.
    Similar to \cref{fig:summary}, the functions $g_i$ are determined by the $i$-th order statistics of $\taskspacetr$, and are explicitly given in  \cref{app:experiment_details}.
    Novel-task error drops below the memorizing oracle as $\phi_3$ is learned ($n \approx 10^5$--$10^6$) and is lowest once $\phi_3$ ($g_1$) saturates.
    The error then rises back toward the memorizing oracle as the network learns higher order statistics through $g_2$ ($n \gtrsim 10^7$).
    Further experimental details are in \cref{app:experiment_details}.
    }
    \label{fig:transience}
\end{figure}

As the training sample size $n$ grows, KR will learn successively lower-eigenvalue (i.e. higher-degree) components of $f_\mathrm{mem}$, on which it is trained.
The kernel will learn the third-order components of $f_\mathrm{mem}$ with approximately $L^2d^2/2$ samples (there are roughly $L^2d^2/2$ quadratic eigenmodes that have larger eigenvalue).
With these third-order components learned, the kernel's function (given a $3$ subscript to denote the order of truncation) becomes approximately
\vspace{-0.7mm}
\begin{equation}
    \hat{f}_{\mathrm{ker}, 3} \approx \hat{\vmu}_\vbeta^\top\vx_{L+1} + \frac{d}{L+d+1}\left(\sum_i y_i\vx_i^\top - \frac{L}{d}\hat{\vmu}_\vbeta^\top\right)\vx_{L+1},
    \label{eq:deg3ker}
\end{equation}
which approximates $f_\mathrm{gen}$, as is seen experimentally in \cref{fig:summary}.\footnote{Just earlier, it was mentioned that the degree $2\ell+1$ term should be defined with the degree $\ell+1$ cumulant, yet the degree $3$ term shown here has no covariance. Why does it not appear? For a resolution, please refer to \cref{subsec:anisotropic,app:correspondence}.}
As $n$ grows further, $\hat{f}$ picks up yet higher-order components of $f_\mathrm{mem}$, eventually arriving at the memorizing solution.
In \cref{fig:transience}, we validate how a model trained without feature learning approximates $f_\textrm{gen}$ when it has learned the low-order eigenfunctions, and begins memorizing using higher-order eigenfunctions.

\section{Transience is an underfitting phenomenon}
\label{subsec:transience}

\cref{sec:eigenstructure} showed that $f_\mathrm{gen}$ is approximately a low-degree truncation of $f_\mathrm{mem}$ (and as we will later see, \cref{subsec:task-space} gives the same picture in the task space).
This is enough to explain the ``transience of ICL'' phenomenon, in which models first generalize and then memorize.

With this simple picture, we are now in a position to resolve the paradox we initially posed: why can a model generalize even though it is optimized solely using the training objective with finite training tasks? %
There is a simple line of logic here.
Models \emph{are always} minimizing the training objective, which is achieved by implementing the memorizing solution.
While minimizing this training objective, they implement a low-degree truncation of the memorizing solution, which is nothing but an approximation of the generalizing solution.

\paragraph{Putting the theory to work.}
While this is a satisfying conclusion, there are still several further questions that we now have the ability to resolve.
In the following sections, we begin by discussing how to map the found eigenfunctions into the more problem-intuitive ``task space'' where the task vectors live.
We then describe why even an optimally truncated model will typically not reach the generalizing solution.
We will similarly analyze the effect of ``task diversity,'' and explain the main findings of \citet{raventos2023pretrainingtaskdiversityemergence}.
Finally, in \cref{sec:transformers}, we show that even transformers, models that do feature learn, implement the same cumulant-truncated solution described by this non-feature-learning theory.
%

% In terms of both eigenfunctions of the data (\cref{sec:eigenstructure}) and, we will later find, the approximated task in task space (\cref{subsec:task-space}), we find $f_\textrm{gen}$ to approximately be a low-degree truncation of $f_\textrm{mem}$. 
% With this fact, we are now well-positioned to understand the ``transience of ICL'' phenomenon, wherein models first generalize before they memorize.

\section{The learning dynamics in task space}
\label{subsec:task-space}

The kernel's eigenfunctions, and therefore computation, are polynomials of the raw data space (i.e. of $\X$ and $\Y$), but most questions of ICL are asked in the more intuitive \textit{task space} $\taskspace$.
Our theory offers a bridge between the two spaces.
All eigenfunctions with non-zero eigencoefficient share one key property allowing us to move between the two spaces---they are linear in $\vx_{L+1}$.
The kernel prediction can thus be written as a linear weight applied to the unpaired input $\vx_{L+1}$: $\hat{f}_\text{ker} = \hat{\vbeta}^\top \vx_{L+1}$ for some estimated task $\hat\vbeta$.
While this estimated task is a function of the random input $\vz$ and not just the underlying task $\vbeta^*$, we can nevertheless predict what the \textit{expected} task a kernel predicts is in terms of only the pretraining task distribution $\d_\mathrm{tr}$, as well as the true task $\vbeta^*$.
We can similarly decompose both $f_\mathrm{gen}$ and $f_\mathrm{mem}$ into a predicted task applied to the unpaired input.\footnote{
This might appear to be unnecessary given the nature of these functions (i.e. $f_\mathrm{gen}$ will clearly predict the underlying task $\vbeta^*$ on average).
The usefulness of this approach is in low-degree truncations of both solutions, similar to the eigenfunction analysis.}
For further information on the task/data-space duality, refer to \cref{app:tasklearning}.
% 

% Although we cannot solve for the eigenfunctions in closed form for every degree, and thus cannot find the eigencoefficients of $f_\mathrm{mem}$ in closed form, there is still regularity in the eigendecomposition of $f_\mathrm{mem}$.
% % 
% As $f_\mathrm{mem}$ is linear in $\vx_{L+1}$, every eigenfunction with a nonzero eigencoefficient must also be linear in $\vx_{L+1}$, the other odd terms having been Gram-Schmidt against the linear term.
% 
% That implies that the kernel prediction itself can be written as a linear function of $\vx_{L+1}$: $\hat{f}_\text{ker} = \hat{\vbeta}^\top \vx_{L+1}$. 
% 
% We can use understand the kernel's predictions not just in terms of the eigenfunctions, but also in the intuitive \emph{task space} $\taskspace$ itself!
% 

\subsection{Evaluating predictors in task space}
\label{subsec:evalpredtaskspace}
Let ${\vbeta}^\top_\mathrm{(\cdot)} \; \vx_{L+1} \equiv \mathbb{E}_{{\vx},\eta}\bigl[ f_{(\cdot)}(\vz)|\vbeta^\star\bigr]$ for $(\cdot)$ being any of $\mathrm{mem}, \mathrm{gen}, \text{or } \mathrm{ker}$ and $\vbeta_{(\cdot)}$ being a function of $\vbeta^*$.
Similar to \cref{sec:eigenstructure}, we can describe $\vbeta_\mathrm{mem}$ and $\vbeta_\mathrm{gen}$ as a sum of terms, thanks to the linearity of the expectation.
%operator.
% 
Each factor of $y_k\vx_k^\top$ evaluates to $\tfrac{L}{d}\vbeta^*$ under the expectation, causing the degree $2\ell+1$ eigenfunctions previously discussed to yield degree $\ell$ functions of $\vbeta*$ (the $+1$ is dropped, being the $\vx_{L+1}$ term).
Before proceeding, we caution that the $2\ell+1$ eigenfunctions $\rightarrow$ $\ell$ functions of $\vbeta^*$ mapping does not always hold.
In particular, we note there exists a degree $5$ eigenfunction with leading term $\tfrac{1}{L^3}(\sum_{i,j} y_i\vx_i^\top (\vx_j\vx_j^\top - \mI_d) - \dots)\vx_{L+1}$, which does indeed appear in the kernel, memorizing, and generalizing solutions.
Many higher order terms appear, although this is learned first and with the greatest coefficient.
As this eigenfunction contains a single $y_i$, it becomes a linear function of $\vbeta^*$ in the task space picture.
(These functions were not previously mentioned as they make only $\mathcal{O}(L^{-1})$ contributions to all functions, and don't greatly impact the overarching narrative.)
Evaluating the per-term generalizing and memorizing predicted tasks, we get,
\begin{align}
    \vbeta_{\mathrm{gen}}(\vbeta^*) &\approx \tfrac{L}{L+d+1}\vbeta^* + \tfrac{(d+1)L}{(L+d+1)(L+d+3)}\vbeta^*+\dots,\\
    \label{eq:taskgenmem}
    \vbeta_{\mathrm{mem}}(\vbeta^*) &\approx \hat{\vmu}_\vbeta + \tfrac{L}{L+d+1}\left(\vbeta^*-\hat{\vmu}_\vbeta\right) + \dots + \tfrac{(d+1)L}{(L+d+1)(L+d+3)}\vbeta^* + \dots.
\end{align}
To leading order in $L$, the low-degree truncation of the memorizing solution's predicted task is still nothing but an approximation of the true underlying task.
A kernel will learn each term of the memorizing weight sequentially: first using just the global mean task $\hat{\vmu}_\vbeta$, then partially forgetting it in favor of approximating the true weight $\vbeta^*$ and generalizing, finally proceeding to compute higher order functions of $\vbeta^*$ to exactly memorize the training tasks.
The dynamical picture in the task space is given in \cref{fig:summary} (left).
\section{What causes even optimally-stopped models to (not) reach $f_\mathrm{gen}$?}
\label{subsec:nogen}

While a model truncated to implement only its degree $3$ eigenfunctions approximates the generalizing solution, there are a number of different effects that can improve the quality of the approximation.
We begin with a discussion of the one present throughout every previous section, the finite context length, before delving into how non-Gaussianity drives memorization.

\subsection{Finite context length $L$}
\label{subsec:finitecontext}

Visible in \cref{eq:taskgenmem,eq:deg3ker}, the degree 3 (data) eigenfunction carries a prefactor $\tfrac{L}{L+d+1}$, after taking into account the summation across context.
That is, models truncated to polynomial degree $3$ capture a $\tfrac{L}{L+d+1}$ fraction of the generalizing solution, which converges rather slowly to $1$ as $L$ grows.
Despite this slow convergence, models that implement this polynomial degree $3$ truncation of $f_\textrm{mem}$ are often achieving their lowest possible loss on novel tasks; implementing higher degree terms will generally raise this loss.
However, in cases where the training tasks are skew-less and do not express the memorizing-only degree $5$ eigenfunction (see \cref{eq:high_phi}), the model is able to capture a further $\tfrac{L(d+1)}{(L+d+1)(L+d+3)}$ fraction of the generalizing solution by learning the (lower-eigenvalue) degree $5$ eigenfunction discussed in \cref{subsec:evalpredtaskspace}.
We conjecture that feature-learning models often learn this generalization-useful term earlier than the naive kernel eigenvalue ordering, when the tasks have a low skewness.

\subsection{Anisotropy in the drawn tasks}
\label{subsec:anisotropic}

\begin{figure}[t]
    \includegraphics[width=0.98\textwidth]{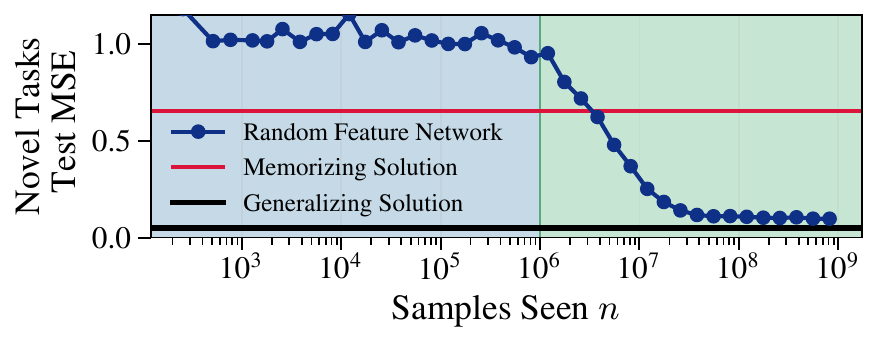}
    \definecolor{phaseoneblue}{HTML}{3058a6}
    \definecolor{phasetwogreen}{HTML}{218152}
    \definecolor{phasethreeorange}{HTML}{bb6719}
    \caption{
        \textbf{Random feature models can perform ICL even with only $T=d+1$ tasks.}
        We train a random feature network using SGD with exactly $T = 4 = d+1$ tasks ($d = 3$), with context length $L=96$.
        We manually construct a training task pool such that $\hat\Sigma = \mI_d$ and $\hat\mu = \boldsymbol{0}$. % and $\sum_a\hat\kappa_{3,aa\_} = \boldsymbol{0}$.
        As is predicted in \cref{subsec:nogen}, this training task distribution allows the model to near-perfectly generalize once the optimal degree-$3$ term $\sum_i y_i \vx_i^\top \vx_{L+1}$ is learned.
        We evaluate the random features model's test error on novel tasks $\beta^\star \sim \d_\mathrm{novel}$, normalized by $\mathbb{E}[y^2]$.
        When the model has not seen many samples in \textcolor{phaseoneblue}{phase one}, the model uses the average training task, which we manually set to $\boldsymbol{0}$.
        At approximately $10^6$ samples, the model enters \textcolor{phasetwogreen}{phase two} and begins fitting the generalizing degree-$3$ term.
        Even though there are only $4$ training tasks, the model does \emph{not} memorize the tasks with up to $10^9$ samples!
        Given infinite training samples, the model will eventually memorize.
        Further details in \cref{app:experiment_details}.
        }
        \label{appfig:ridge_reach}
\end{figure}

While we have been using $\phi_3 \sim \tfrac{d}{L+d+1}(\sum_i y_i\vx_i^\top - \tfrac{L}{d}\hat\vmu_\vbeta)\vx_{L+1}$ throughout to denote \emph{the} ``generalizing'' eigenfunction, this is only a true eigenfunction when the training tasks have an isotropic covariance.
As a very simple demonstration of this effect, we train a network on a $d=3$ dimension setup with only $T=4$ tasks carefully selected to have isotropic covariance.
With only $4$ tasks, the lazy model is able to generalize, as is shown in \cref{appfig:ridge_reach}.
The accurate ``generalizing'' eigenfunctions therefore are defined using the training task's covariance (specifically its singular value decomposition), and any anisotropies will induce error in the novel task prediction.
For a complete description of the eigenfunctions, refer to \cref{apptable:eigcoeffs}.

\subsubsection{Anisotropic inputs}
\label{subsubsec:aniso-xs}

While anisotropy in the sampled tasks affects downstream performance, we can also ask how sampling anisotropic inputs $\vx$ affect the learned solution.
We point to \cref{app:correspondence} for a complete description, but can note that $\vbeta^\top\vx_i = \left(\vbeta^\top \Sigma_\vx^{1/2}\right)\left(\Sigma_\vx^{1/2}\vx_i\right)$, meaning anisotropic inputs can be viewed as whitened inputs with the tasks themselves being colored.

\subsection{Task Diversity/Non-Gaussianity of Tasks}
\label{subsec:taskdiv}

\begin{figure}[t!]
    \includegraphics[width=\textwidth]{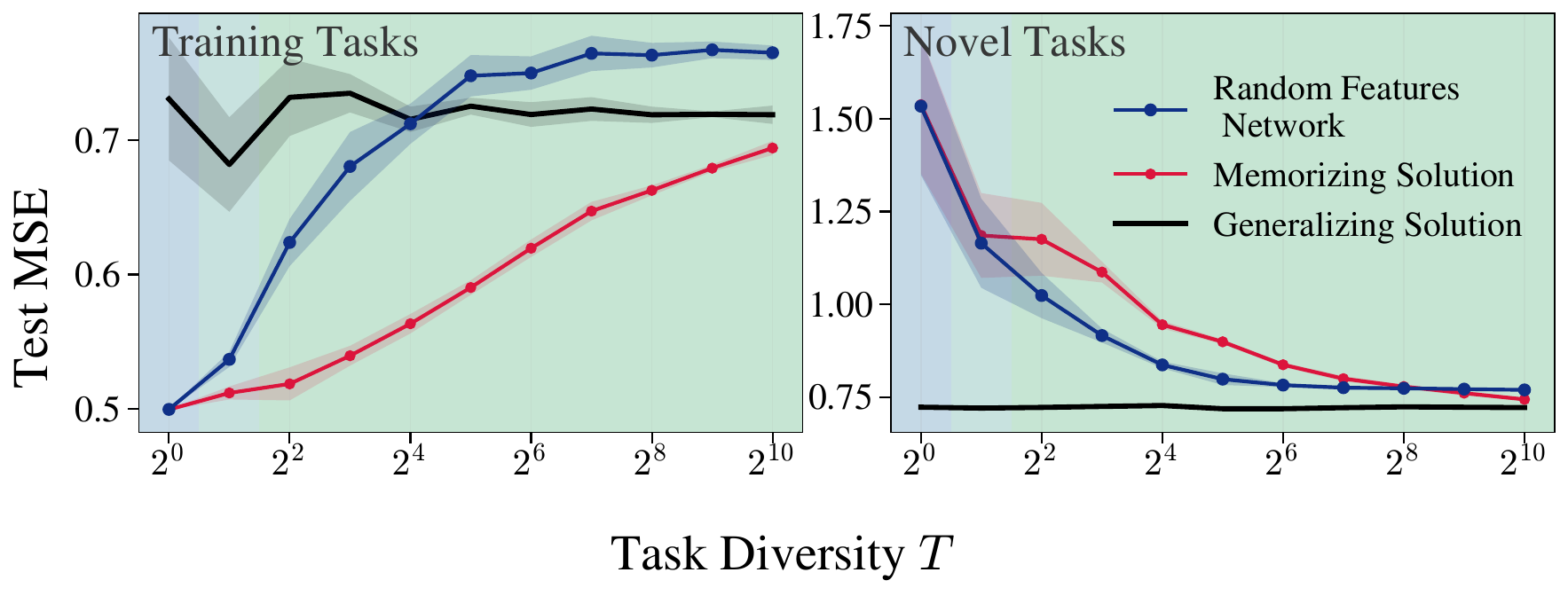}
    \caption{
    \definecolor{phaseoneblue}{HTML}{3058a6}
    \definecolor{phaseonefivebleen}{HTML}{B4DAD6}
    \definecolor{phasetwogreen}{HTML}{218152}
    \textbf{Task diversity trades pretraining-task accuracy for novel-task accuracy.}
    Following \citet{raventos2023pretrainingtaskdiversityemergence}, we sweep the number of training tasks $T$ and report test MSE on the pretraining tasks $\d_\mathrm{tr}$ (\textbf{left}) and on novel tasks $\d_\mathrm{novel}$ (\textbf{right}).
    Our data is drawn with $d=8, L=15, \epsilon^2=0.25$.
    In blue, we show a fit random features network trained with SGD (see \cref{app:experiment_details} for random feature network details), with the memorizing (\textcolor{red}{dMMSE}) and generalizing (\textbf{ridge}) oracles shown.
    As $T$ grows, the training task distribution's cumulants approach that of the prior's (Gaussian) and all three curves converge.
    Evaluated on novel tasks, the random features model initially tracks the memorizing solution up to a task diversity $T$ of $4$, continuing to approach the generalizing solution until $T\approx 2^6=64$ tasks.
    All models are trained with enough samples such that they can fully realize the optimal degree-$3$ term, indicated by the \textcolor{phasetwogreen}{green} background.
    We note two exceptions: a \textcolor{phaseoneblue}{blue} background for one training task is used to signify that the model uses the mean task alone, and the \textcolor{phaseonefivebleen}{seafoam} background for two tasks signifies the mean task still being heavily utilized.
    The trained models do not exactly reach the generalizing solution, in part due to the finite context length $L$ we use for training (see \cref{subsec:nogen}).
    Further experimental details are in \cref{app:experiment_details}.
    }
    \label{fig:kernel_sweep}
\end{figure}

If models learn the generalizing solution by learning only low-degree functions of the memorizing solution---those defined by the Gaussian statistics of the training tasks---how should we understand the task diversity picture of \citep{raventos2023pretrainingtaskdiversityemergence}?
In their work, providing a transformer more task diversity caused generalization, and transformers given fewer tasks would memorize.
The story of task diversity is actually independent of the transformer architecture (as was hinted in \citet{tong2025mlpslearnincontextregression}), not even needing feature learning to obtain the headline results (see \cref{fig:kernel_sweep}).
As task diversity increases, models begin generalizing in large part due to the training task statistics becoming both more Gaussian, and more isotropic.
As the tasks are drawn from an isotropic distribution, more tasks causes the sampled covariance to become increasingly white.\footnote{The eigenvalues of $\hat\Sigma_{\vbeta}$ follow the Marchenko--Pastur distribution, with increasing task diversity moving the eigenvalues closer to $\Sigma_\vbeta$, as was first described in the ICL setting by \citet{letey2025pretraintesttaskalignmentgoverns}.}
Furthermore, drawing more tasks causes the higher order cumulants to decrease, decreasing the scale of the memorization-only eigencoefficients.
Both of these effects are further discussed in \cref{appsubsec:dmmsevsridge}.
\section{Transformers learn $f_\mathrm{gen}$ using Gaussian statistics}
\label{sec:transformers}

\begin{figure}[t!]
    \includegraphics[width=\textwidth]{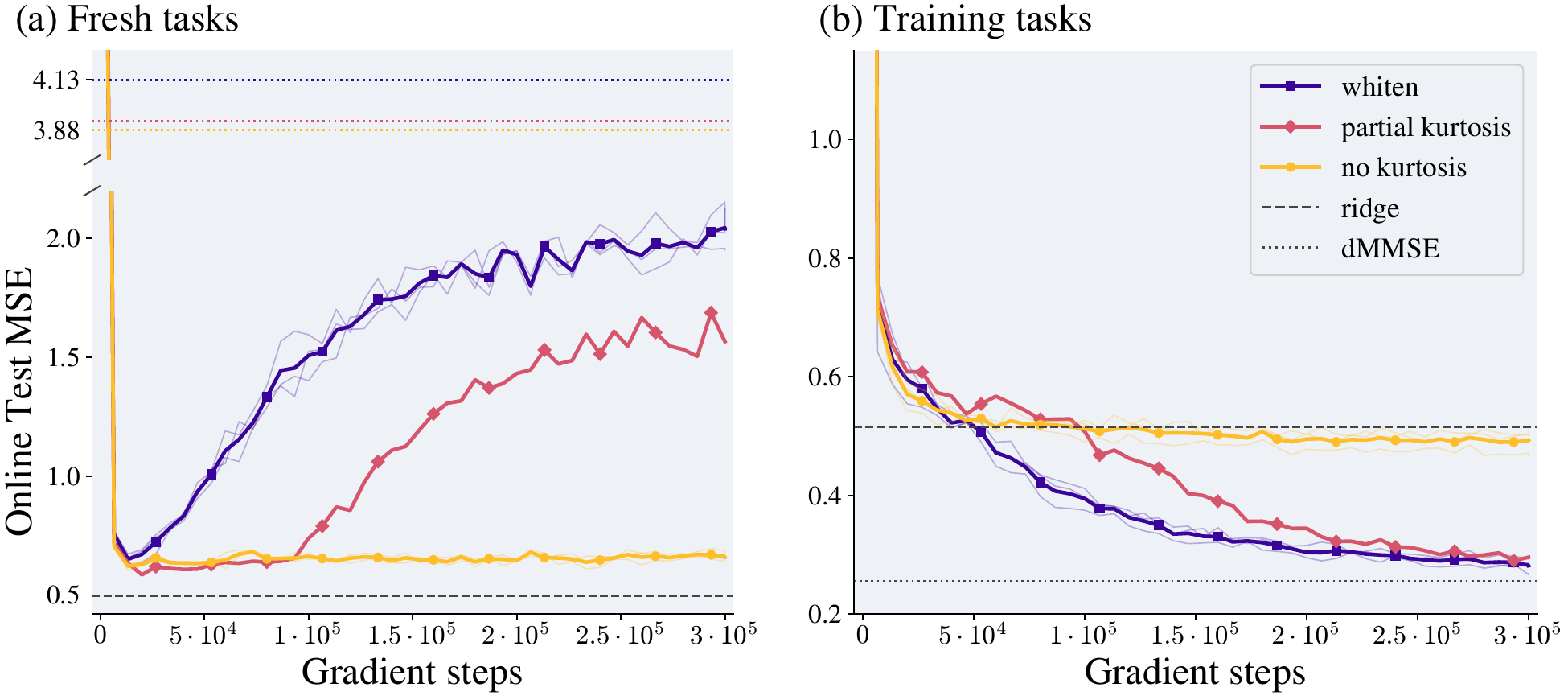}
    \caption{
    \textbf{Transformers memorize through non-Gaussian statistics of their training tasks.}
    We train $12$-layer GPT-2-style transformers with Adam on $T=256$ training tasks ($d=8$, $L=16$, $\epsilon^2=0.25$) and evaluate the test MSE on \textbf{(a)} fresh tasks and \textbf{(b)} the training tasks.
    All pools are centered and whitened (see \cref{subsec:anisotropic} for a discussion of the implications) and differ principally in their fourth cumulant.
    \textit{Whiten} consists of i.i.d.\ Gaussian draws,
    \textit{partial kurtosis} is made of a symmetric pool whose $\hat\kappa_4$ is rescaled to $0.3\times$ that of its initial Gaussian pool, \textit{no kurtosis} uses a symmetric pool with $\hat\kappa_4=0$.
    Dashed and dotted lines mark the generalizing (ridge) and memorizing (dMMSE).
    In \textbf{(a)}, each pool's dMMSE is shown to highlight that the slight difference in oracle does not explain the onset of task memorization.
    All models first approach the generalizing solution, with the scale of the kurtosis setting when memorization begins.
    Further experimental details are in \cref{app:experiment_details}.
    }
    % \caption{
    % \definecolor{phaseoneblue}{HTML}{3058a6}
    % \definecolor{phaseonefivebleen}{HTML}{B4DAD6}
    % \definecolor{phasetwogreen}{HTML}{218152}
    % \textbf{Transformers learn the generalizing solution using cumulants}
    % BAD CAPTION, CHANGE
    % % \remind
    
    % }
    \label{fig:transformer}
\end{figure}

We now arrive at the often uncomfortable question: %
having done theory on models unable to do feature learning, can we say anything about those that have adaptive features?
We find the answer here is yes: even transformer models trained with the Adam optimizer learn to memorize by exploiting the non-Gaussian statistics of their training tasks.\footnote{It is also well-appreciated in the literature (see e.g. \citet{nakkiran2019sgdneuralnetworkslearns}) that neural networks learn simple statistics first.}
Shown in \cref{fig:transformer}, we fix a total number of tasks, and train transformers on datasets containing various amounts of training task kurtoses (having been drawn symmetrically such that there is no skewness).
When the model is given no kurtosis, it remains generalizing throughout the entire training procedure.
When the model is given only part of the full kurtosis, it will stay generalizing longer than a model given a typical kurtosis.
This is the ordering our theory predicts: %
with all odd cumulants removed by the symmetric pool, the first memorizing term is the degree-$7$ mode carrying $\hat{\kappa}_4$ (\cref{thm:cum-aniso}).
Removing $\hat{\kappa}_4$ leaves only $\hat{\kappa}_{\ge 5}$ to memorize, which occurs much later in training.
% 

% For models that implement simple functions early in training (), we can attribute transience to these models learning the low-degree truncation of the memorizing solution as opposed to implementing the generalizing solution.
% 
% Even in our setting where we train models that do not feature learn, we still predict transience for kernels in \cref{fig:summary} and observe it in random feature models (which do not feature-learn) in \cref{fig:transience}.
% 
% \input{sections/bayes_oracles}
% \input{sections/learning_curves}
\section{Further Discussion}
\label{sec:discussion}

Our analysis is restricted to the in-context linear regression task with Gaussian inputs.
While we believe the generalizing solution is contained within low-degree functions of the memorizing solution for ICL settings, and is thus the cause of generalization in ICL, we do not attempt to show this in full generality.
We leave future work to characterize the effects of learning nonlinear ICL solutions or how natural-language affects the learned function.
Furthermore, the memorization phase is particularly difficult to reach for models unable to perform feature learning; %
despite this, \cref{fig:transformer} shows that our account of memorization---sequentially learning higher order statistics of the training tasks---holds even in transformer models.

In closing, we remind that our results were derived from simple, solvable models, and yet have provided insight into an aspect of transformers performing a seemingly complex task.
We therefore hope this work shows the usefulness of solvable models, even ones that might initially seem ineffective for the problem at hand, to advance our understanding of nebulous phenomena.

\subsubsection*{Acknowledgments}
We are grateful to Dhruva Karkada, Blake Bordelon, Michael DeWeese, Sultan Daniels, Divit Rawal, and Berkan Ottlik for useful discussion on key results.
We further thank Dhruva Karkada, Berkan Ottlik, Raymond Fan, Samyak Jain, and Mark Rhee for providing feedback on the manuscript.
In addition to the above, JT thanks Amy Flather and Soren Bear for listening to his early rants about why ICL is important and should be studied using kernels, as well as Josh Albrecht and Kanjun Qiu for continuing to support his fundamental AI research efforts.

This work was funded by Imbue under the Feature Lab (FLAB) initiative.

\subsubsection*{Statement on the Use of Large Language Models}
We used large language models (LLMs) for writing code, giving a precise expansion of the Ridge and dMMSE functions in terms of cumulants and drafting their appendices, refining the mathematics of high order subkernel diagonalization, helping clarify the position of related works, and constructing proofs.
As LLMs often made various implicit assumptions about the structure of the empirical cumulants, we independently verified all outputs of LLMs.

\newpage

\let\oldurl\url
\def\url#1{}
\bibliography{refs}
\bibliographystyle{style/iclr2027_conference}
\let\url\oldurl

\newpage
\appendix
% \tableofcontents
\section{Review of Notation}
\label{app:notation}

As notation in ICL literature evolves constantly, we provide a table to reference all important quantities.

% \begin{table}[h]
%     \centering
%     \renewcommand{\arraystretch}{1.1}
%     \begin{tabular}{c|c}
%         \remind Symbol & Meaning\\
%         \hline
%         $d$ & In-context data dimension\\
%         $L$ & Context length\\
%         $T$ & Number of pretraining tasks \\
%         $\vbeta$ & General notation for tasks\\
%         $\mathcal{B} = \{\vbeta^{(t)}\}_{t=1}^T$ & Pool of tasks\\
%     \end{tabular}
%     % \caption{Quantities appearing throughout the main document}
% \end{table}

\begin{table}[ht]
    \centering
    \renewcommand{\arraystretch}{1.15}
    \begin{tabular}{c|l}
        Symbol & Meaning\\
        \hline
        \multicolumn{2}{c}{\textit{Data and task distribution}}\\
        \hline
        $d$ & In-context data dimension\\
        $L$ & Context length\\
        $T$ & Number of pretraining tasks\\
        $n$ & Number of training samples (contexts) supplied to the kernel\\
        $\vbeta$ & General notation for tasks\\
        $\taskspaceprior = \mathcal{N}(\mu_\vbeta, \Sigma_\vbeta)$ & Prior distribution used to draw tasks\\
        $\taskspacetr = \{\vbeta^{(t)}\}_{t=1}^T$ & Training pool of tasks\\
        $\vz$ & Context (samples) consisting of $(\vx_1^\top,y_1,\dots,\vx_L^\top,y_L,\vx_{L+1}^\top)$\\
        $\Sigma_\vx$ & Covariance of the in-context data $\vx_i$\\
        $\epsilon^2$ & Label noise variance, $\eta_i \sim \mathcal{N}(0, \epsilon^2)$\\
        $\delta$ & Ridge parameter in KRR\\
        $\hat{\vmu}_{\vbeta},\ \hat\Sigma_{\vbeta}$ & Mean and covariance of $\d_\mathrm{tr}$ (i.e.\ $\hat\kappa_1,\hat\kappa_2$)\\
        $\kappa_\ell$, $\hat\kappa_\ell$ & Order-$\ell$ cumulant of $\d_\mathrm{novel}$ and $\d_\mathrm{tr}$, respectively\\
        $M_m$ & Order-$m$ raw moment tensor of $\d_\mathrm{tr}$\\
        \hline
        \multicolumn{2}{c}{\textit{Whitened quantities} (Given $\Sigma_\vx \neq cI_d$)}\\
        \hline
        $\tilde\vx = \Sigma_\vx^{-1/2}\vx$ & Whitened in-context data\\
        $\tilde\vbeta = \Sigma_\vx^{1/2}\vbeta$ & Colored task; $\tilde{\mathcal{B}} = \Sigma_x^{1/2}\mathcal{B}$\\
        $\tilde\kappa_\ell$ & Cumulants of $\tilde{\mathcal{B}}$\\
        \hline
        \multicolumn{2}{c}{\textit{Kernel and eigenstructure}}\\
        \hline
        $c_\ell$ & Level coefficients of an inner-product kernel (\cref{def:inner-product-kernel})\\
        $K_\ell$ & Degree-$\ell$ subkernel (\cref{def:inner-product-kernel})\\
        $\lambda_i,\ \phi_i$ & Kernel eigenvalues and (orthonormal) eigenfunctions\\
        $v_i$ & Eigencoefficient, $f^\star = \sum_i v_i\phi_i$\\
        $\mathcal{L}_i$ & Learnability of mode $i$; $\hat f = \sum_i \mathcal{L}_i v_i\phi_i$\\
        $\zeta$ & Eigenframework constant solving Eq.~\eqref{eqn:kappa_def_eqn_app}\\
        $\mathcal{E}_0$ & Overfitting coefficient\\
        $C_{FF},C_{FL},C_{LL},\tilde C_\ell$ & Level covariance blocks and projected covariance of \cref{def:see}\\
        $S_{ij},\ \tilde v_i$ & Eigenfunction overlap and test eigencoefficients on $\d_\mathrm{novel}$\\
        % \hline
        % \multicolumn{2}{c}{\textit{Channel grading} (\cref{thm:cum-aniso})}\\
        % \hline
        % $\omega$ & Cumulant order carried by a channel; one plus the number of $y$'s\\
        % $k$ & Number of self-contracted label pairs ($y_\ell^2$ factors)\\
        % $\Tr^k[\kappa_\omega]$ & $\kappa_\lambda$ with $k$ disjoint index pairs traced\\
        % $G_{\omega,k},\ H_{\omega,k}$ & Across-pool and within-context Gram terms\\
        % $c_{\omega,k}$ & Scalar prefactor $(1 + H/LG)^{-1/2}$ when $H\propto G$\\
    \end{tabular}
    \caption{Quantities appearing throughout the main document and appendices.}
\end{table}

% Before going into the derivation, let's define a number of terms that will be useful going forward:
% % 
% Let $S_\alpha \equiv \sum_{i=1}^L\vx_{i,\alpha}y_i$, and $\mS = [S_1, \ldots, S_d]^\top$.
% % 
% Orthogonalized against the linear term, we set
% \begin{equation}
%     \phi_\ag = \phi_\ag^\perp + L(\Sigma_x\bar\vbeta)_\alpha\cdot \vx_{L+1, \gamma}.
% \end{equation}
% % 
% As we are solving for general $\vx$-covariance of $\Sigma_x$, it will be useful to introduce whitened parameters, always defined with a tilde:
% \begin{align}
%     \tilde{\vx}_{L+1} &= \wS \vx_{L+1}\\
%     \tilde{\mS}_\alpha &= \wS \mS_\alpha\\
% \end{align}

\section{Experimental Details}
\label{app:experiment_details}

In \cref{fig:summary,fig:transience}, we introduced $g_i$ to describe polynomial functions learned by the kernel.
The functions $g_i$ are chosen to be representative of the polynomials of degree $2i+1$ with large eigencoefficients: that is, $g_0$ was randomly selected as one of the $d$ eigenfunctions $\vx_{L+1}$, $g_1$ was chosen to be eigenfunction most closely related to $\phi_3(\vz) \sim \left(\sum_iy_i\vx_i^\top -\tfrac{L}{d}\hat\vmu_\vbeta^\top\right)\vx_{L+1}$, $g_2$ was one of the $d$ eigenfunctions most closely related to (at leading order) $\vphi_5 \sim \sum_{ij} y_iy_j\vx_i^\top\vx_j\vx_{L+1}$, and finally $g_3$ was the eigenfunction most closely related to (at leading order) $\phi_7 \sim\sum_{ijk} y_iy_jy_k\vx_i^\top\vx_j\vx_k^\top\vx_{L+1}$.

\subsection{Exact experimental setups}
\label{appsubsec:analyticalexpts}

We ran three main types of experiments: exact kernel experiments, exact random feature solvers, and random feature (RF) model solutions found via SGD.
In the main text, we detail only the kernel experiments and RF model experiments, as those were the most fruitful.
Below, we describe the setup for each figure in the main text, along with our rationale for using random features models in place of kernels.

\paragraph{The ICL Kernel.}
A kernel method solving the ICL linear regression task requires an enormous number of samples to get to either of the Bayes-optimal predictors.
We can, however, reduce this by very carefully choosing a kernel function constructed to remove as many terms (those with eigencoefficient $0$) as possible.
We use the subkernel expansion framework (\cref{def:see}) and the eigencoefficients found in \cref{app:correspondence}, to make the ``ICL Kernel'', a Gaussian kernel with the linear components of $\vx_{L+1}$ extracted, and the even polynomials removed:
\begin{equation}
    K_{ICL}(\vz, \vz') = \Big(\exp(-|\vc-\vc'|^2/2\sigma^2)+\exp(-|\vc+\vc'|^2/2\sigma^2)\Big)\vx_{L+1}^\top \vx_{L+1}',
    \label{eq:iclkernel}
\end{equation}
where $\vc$ is nothing but $\vz$ with the unpaired datapoint $\vx_{L+1}$ removed.
This ``ICL Kernel'' that is often used in the experiments is not truly rotation invariant due to the $\vx_{L+1}$ term; nevertheless, it contains the same eigenfunctions with non-zero eigencoefficients as the Gaussian kernel (shown empirically to match the theory in \cref{fig:eigendecom,fig:learning_curves}).
Therefore, we readily use this ``ICL Kernel'' in experiments and make the informed guess that rotation-invariant kernels would behave identically, given sufficient data.

\paragraph{Kernel model experiments.}

In the right half of \cref{fig:summary}, we train an ICL kernel with kernel width $\sigma =3$, a ridge parameter $\delta = 10^{-2}$, and each individual sample partially whitened such that $\sum_i \vx_i \vx_i^\top = I_d/d$.
We take $d=5$, $L=7$, and $\epsilon^2 = 0.1$.
The in-context data is sampled $\vx_i \sim \mathcal{N}(0, I_d/(dL))$.
We draw $T=16$ tasks to get $\taskspacetr$, taken from a prior of $\mathcal{N}(0, I_d)$.
For the trained kernel models, we use $20$ trials, keeping $\taskspacetr$ fixed.

\paragraph{Direct eigenstructure checks.}
\label{apppara:eigenstructure-comp}
In \cref{fig:eigendecom}, we check that the subkernel expansion framework (\cref{def:see}) accurately predicts the eigendecomposition of a trained kernel.
The two spectra are straightforward to compare: we sort both in decreasing order and overlay them against mode index on log--log axes.
Agreement means the curves coincide.
The eigenbases are more difficult, most notably due to the ordering.
The spectrum is degenerate over many different sections, and so assigning a cosine similarity between a theoretical eigenvector and empirical one mode-by-mode is at best left to chance.
Therefore, we must bin the eigenvectors and compare their overall overlap.

We instead place bin edges in the gaps of the spectrum between degenerate sectors at the geometric mean
$\sqrt{\lambda_{(m)}\lambda_{(m+1)}}$ of each consecutive pair whose ratio exceeds a hyperparameter threshold $r$.
We additionally consolidate many bins at smaller eigenvalues (defined by another hyperparameter) into 2 bins such that our images are not dominated by many bins that are empirically noisy.

Let $d^{(\mathrm{emp})}_\ell$ be the number of empirical modes in bin $\ell$.
Let $\Phi^{(\mathrm{emp})}_\ell \in \mathbb{R}^{n \times d^{(\mathrm{emp})}_\ell}$ collect the empirical eigenvectors of bin $\ell$ as columns, and let $\Phi_k$ do the same for the predicted eigenfunctions of bin $k$.
The pairwise subspace overlap is
\begin{equation}
    \mathrm{O}(k,l) = \frac{1}{d^{(\mathrm{emp})}_l}\left\|\,\Phi_k^{\top}\Phi^{(\mathrm{emp})}_l\right\|_F^2.
    \label{eq:overlap}
\end{equation}

The bottom plots of\cref{fig:eigendecom} shows $\mathrm{O}(k,l)$ as a heatmap with eigenvalue increasing towards the top left, so that the horizontal axis runs in the same direction as mode index in the panel above.

We use $n = 25600$ samples.
Our setup consistently uses $d=4$, $\epsilon^2=0.1$, $\Sigma_\vx = I_d/(dL)$, and $\Sigma_\vbeta = I_d$.
All setups use $L=6$, outside of the Laplace kernel which uses $L=9$ such that the inner-product kernel expansion remains valid (concentration of norm of $\|Z\|$), as is described in more detail in B.3 of \citep{karkada2026predictingkernelregressionlearning}.
Outside of the Laplace kernel for norm concentration reasons, all kernel:task-diversity pairs were chosen at random.
In order of their appearance:
\begin{itemize}
    \item T=1, Inner-product kernel uses level coefficients $c_0=1$, $c_1=1/9$, $c_2=1/81$, $c_3=1/729$, with bin edges of $r=1.01$, and the coarse graining of the tail beginning at $\lambda=10^{-5}$, ending at $\lambda=10^{-7}$.
    \item T=4, ReLU NTK uses bias and weight variances of $1$, bin edges of $r=1.018$, and the coarse graining of the tail beginning at $\lambda=7\cdot10^{-5}$, ending at $\lambda=6\cdot10^{-6}$.
    \item T=16, Gaussian kernel uses kernel width of $3$, with bin edges of $r=1.008$, and the coarse graining of the tail beginning at $\lambda=10^{-4}$, ending at $\lambda=10^{-6}$.
    \item T=256, Laplace kernel uses kernel width of $3$, with bin edges of $r=1.01$, and the coarse graining of the tail beginning at $\lambda=8\cdot10^{-5}$, ending at $\lambda=10^{-5}$.
\end{itemize}

\paragraph{Learning curves.}
\label{apppara:learncurve_exptdetails}
In \textbf{\cref{fig:learning_curves}}, we train a Gaussian kernel with kernel width parameter $\sigma = \sqrt3$ and ridge $\delta = 0.1$, on isotropic ICL data with $d=5$ intrinsic dimensions, context length $L=7$, and noise variance $\epsilon^2=0.1$.
Both in-context data $\vx$ and the tasks $\mathcal{B}$ are drawn isotropically as $\vx \sim \mathcal{N}(0, I_d/(Ld))$ and $\vbeta \sim \mathcal{N}(0, I_d)$.
All runs are performed across $20$ seeds with the max $n_\text{samples}$ set by the memory constraints of a Google Colab A-100 GPU: roughly 25,600 samples.
We sweep over $4$ (roughly logspaced) values of task diversity T: $1, 4, 16, 256$ to show our theoretical eigensystem is correct regardless of task diversity and regardless of what distribution we evaluate our theory on.

\subsection{Random Feature Analytical Solver}
\label{appsubsec:rf_analytical}

\begin{figure}[ht]
    \includegraphics[width=\textwidth]{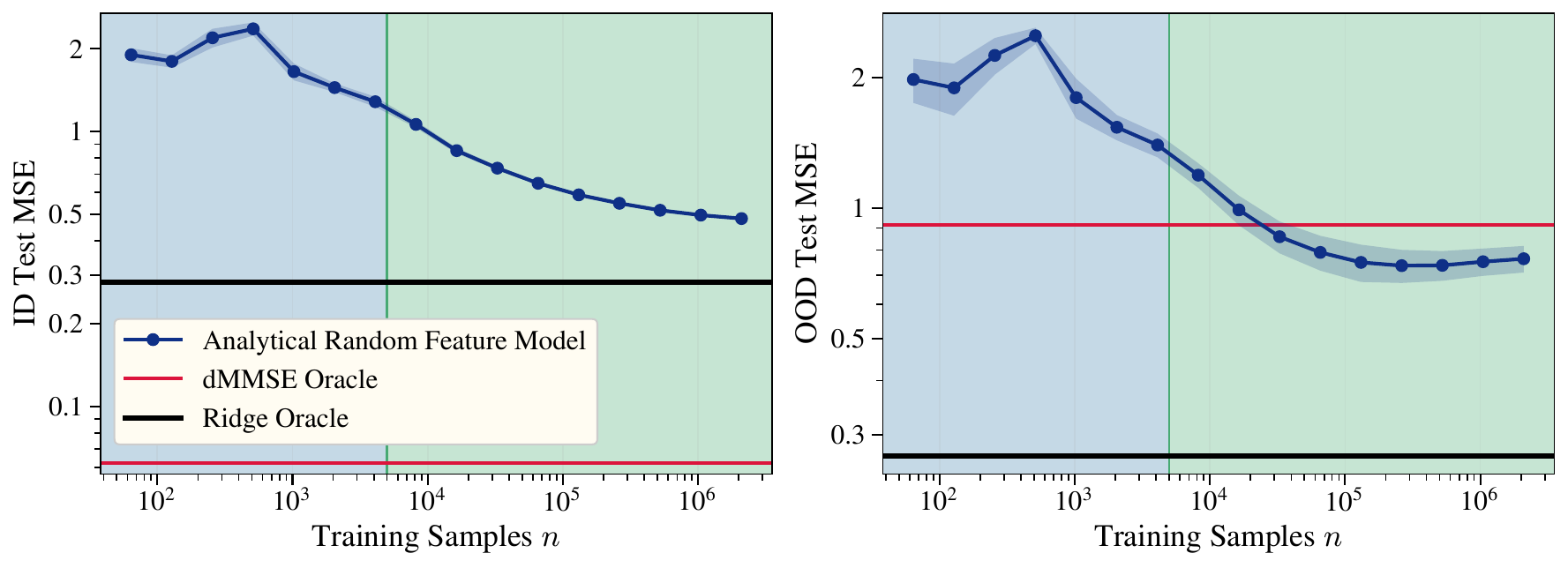}
    \caption{
        \textbf{Analytically solved random feature models outperform the memorizing oracle on novel tasks.}
        We show test MSE on novel tasks ($\d_\mathrm{novel}$) and on the training tasks $\vbeta \in \d_\mathrm{tr}$ of a random feature model whose readout layer weight has been analytically solved, as a function of the number of training samples $n$.
        The limiting kernel function is the ICL kernel of \cref{eq:iclkernel}, set with kernel width $\sigma = 3$.
        We use $T=8$, $L=4$, $d=4$, $\epsilon^2=0.05$, and tasks drawn from an isotropic prior.
        Shading shows the standard deviation over 5 trials.
        Past $n\approx 4\times10^3$, the model's error drops below that of the memorizing oracle (\textcolor{red}{dMMSE}).
        It remains well above the generalizing oracle (\textbf{ridge}) throughout.
        }
        \label{appfig:analytic_rf}
\end{figure}

As exact kernel solvers scaled poorly with the number of samples (one needs to invert an $n \times n$ matrix), random features models were additionally considered.
Kernel models with suitable kernels can be understood as an infinite-width two-layer MLP with the first layer frozen (i.e. a fixed feature map followed by a linear projection) \citep{rahimi:2007}.
Random features (RF) models serve as their finite-width counterpart, so long as the random features defining the frozen layer give the correct NNGP in the limit that the number of features approaches $\infty$.
RF models have a computational advantage: with $p$ random features, fitting the readout by ridge regression only requires solving a $p \times p$ linear system, reducing the training cost from $\mathcal{O}(n^3)$ time and $\mathcal{O}(n^2)$ memory to $\mathcal{O}(np^2 + p^3)$ time and $\mathcal{O}(np)$ memory.

So long as the model is not limited by the number of features (a feature-limited regime), extra performance can be gained from using RF models.
To see the models are in a \textit{data-limited} regime, one can note that they are mostly noise-dominated: %
the non-zero eigencoefficients are sparsely populating our spectrum, and worse than that, the eigencoefficients are not particularly contained within the large eigenvalue region (the largest eigencoefficients appear at the degree-$3$ part of the spectrum corresponding to $\sum_i y_i\vx_i^\top\vx_{L+1}$, for which we have to fit the noisy degree $1$ and $2$ eigenfunctions).
Therefore, we use RF models to substantiate our claims, directly solving for the optimal readout layer.
This culminates in \cref{appfig:analytic_rf}, where, although more samples are able to be fit, it nevertheless remains difficult to see our theory's predicted characteristics (though the drop and rise still are present).
We therefore need even more samples to be fit, suggesting that we move from analytical expressions to ones obtained using SGD.

\subsection{Random Feature Experiments}
\label{appsubsec:rf_expts}
 
As previously mentioned, the RF model is nothing but a special type of two-layer MLP, and training said MLP with online SGD would allow us to trade memory constraints for runtime.

In \cref{fig:transience}, we train an RF model using SGD using $3.3\cdot10^6$ gradient steps with a batch size of $1024$.
The dataset uses $d=4,L=5,\epsilon^2=0.05$.
Our RF models use random features drawn such that they converge to the ICL kernel of \cref{eq:iclkernel}.
The kernel width parameter used for the converging kernel was chosen as $\sigma = 1.1\|\vz\|$.
We draw $T=8$ tasks $\mathcal{N}(0, \mI_d)$.
The tasks are then whitened such that the covariance is exactly $\mI_d$, however we keep the finite non-zero task mean.
Our data is drawn $\vx \sim \mathcal{N}(0, I_d/d)$.
We normalize our labels $y$ by $d\sigma_x^2+\epsilon^2$.
We run our experiments and average across 3 trials, shading the standard deviation.
We used only $8192$ features for our model.
We use a learning rate $\eta = 0.75\cdot\tfrac{2}{\lambda_1}$, $\lambda_1$ being a numerically estimated maximum eigenvalue of the Hessian found via power iteration.
% To reduce random noise, we weight average the random feature model weights of at most $20,000$ steps, and exit our training loop prematurely once we are within a tolerance of $0.01$ of the predicted optimal polynomial degree $3$ solution.

In \cref{appfig:ridge_reach}, we train a RF model on $d=3$ dimensional data with \emph{only} $T=4$ tasks.
We use the same ICL kernel of \cref{eq:iclkernel} as our limiting kernel for which the features are drawn.
The kernel width parameter used for the converging kernel was chosen as $\sigma = 1.5\|\vz\|$.
These tasks are initially drawn from a white Gaussian prior, but are manually whitened and centered: $\mathrm{Cov}_{\mathcal{B}_\mathrm{tr}}=\mI_d$, and the mean training task is set to $\mathbf{0}$.
We take a large context length of $L=96$ and $\epsilon^2 = 0.05$.
We draw data $\vx\sim\mathcal{N}(0, I_d/d)$.
We perform only 1 trial.
We normalize our labels $y$ by $d\sigma_x^2+\epsilon^2=1.05$ to keep the scale invariant to choice of $\Sigma_x$.
We use $65,536$ features for our model.
We train for $3.2\cdot10^6$ gradient steps with a batch size of $256$.
We use a learning rate $\eta = 0.8\cdot\tfrac{2}{\lambda_1}$, $\lambda_1$ being a numerically estimated maximum eigenvalue of the Hessian found via power iteration.

In \cref{fig:kernel_sweep}, we train an RF model using SGD using a maximum of $4\cdot10^6$ gradient steps with a batch size of $1024$.
We exit our training loop prematurely once we are within a tolerance of $0.01$ of the predicted optimal polynomial degree $3$ solution's novel test error (to save on compute time; models would need to train on at least two orders of magnitude more samples before the degree $5$ polynomials begin to be learned).
To reduce random noise, we weight average the random feature model weights of at most $20,000$ steps.
The dataset uses $d=8,L=15,\epsilon^2=0.25$.
Our RF models use random features drawn such that they converge to the ICL kernel of \cref{eq:iclkernel}.
The kernel width parameter used for the converging kernel was chosen as $\sigma = 1.5\|\vz\|$.
Our data is drawn $\vx \sim \mathcal{N}(0, I_d/(d\cdot4))$, the numerical factor of $4$ added such that $d\sigma_x^2=\epsilon^2$ (which we found best portrayed our theory).
We sweep the task diversity $T$ from $2^0$ to $2^{10}$, training a new model per task diversity and run.
We normalize our labels $y$ by $d\sigma_x^2+\epsilon^2=1/2$ to keep the scale invariant to choice of $\Sigma_x$.
We run our experiments and average across 3 trials, shading the standard deviation.
We used $16,384$ features for our model.
We use a learning rate $\eta = 0.8\cdot\tfrac{2}{\lambda_1}$, $\lambda_1$ being a numerically estimated maximum eigenvalue of the Hessian found via power iteration.

\subsection{Transformer Experimental Details}
\label{appsubsec:transformer_experiment}

We follow the setup of \citet{raventos2023pretrainingtaskdiversityemergence}.
We train a GPT-2-style decoder model online with 12 transformer blocks of width $256$, $8$ attention heads, MLPs of width $1024$, GELU activations, no dropout, a causal attention mask, and learned positional embeddings initialized from $\mathcal{N}(0,0.02^2)$.
We train using Adam (PyTorch defaults $\beta_1=0.9$, $\beta_2=0.999$) with a constant
learning rate of $10^{-4}$ and batch size $64$.
There is no weight decay, warm-up or gradient clipping.
Models are trained for $3\cdot10^5$ steps.
The dataset uses $d=8,L=16,\epsilon^2=0.25$.
The in-context data is drawn $\vx_i\sim\mathcal{N}(0,\mI_d)$.
During training, $\beta$ is drawn uniformly from a fixed pool $B_{\mathrm{tr}}$ of $T=256$ tasks. 

Every pool satisfies $\hat\kappa_1=0$ and $\hat\kappa_2=I_d$ exactly, via ZCA whitening.
For the individual trials,
\begin{itemize}
  \item \emph{Whiten:} $T$ i.i.d.\ draws from $\mathcal{N}(0,\mI_d)$, centered and
  ZCA-whitened. The higher cumulants are left at their random sampling: $\|\hat\kappa_3\|_F\approx1.6$ and $\|\hat\kappa_4\|_F\in[5.2,5.7]$ across
  seeds.
  \item \emph{Symmetric pools:} we optimize $T/2=128$ vectors $\vw_t$ and take the pool to be the ZCA-whitened set $\{\vw_t\}\cup\{-\vw_t\}$, so every odd cumulant
  vanishes.
  Starting from $\vw_t\sim\mathcal{N}(0,\mI_d)$, we run L-BFGS (strong-Wolfe line search, at most $2000$ iterations) on
  \begin{equation*}
    \mathcal{L}'=\frac{1000}{24} \mathbb{E}_\vu \left[\big(k_4(\vu)-\alpha k_4^{(0)}(\vu)\big)^2\right] +\frac{1}{720} \mathbb{E}_\vu \left[k_6(\vu)^2\right],
  \end{equation*}
  where $\vu$ ranges over $6000$ fixed random unit vectors and $k_\ell(\vu)$ are the cumulants of the projections $\vu^\top\vbeta$ over the pool:
  $k_4=m_4-3$ and $k_6=m_6-15m_4-10m_3^2+30$ (here $m_3=0$).
  $k_4^{(0)}(u)$ is the directional kurtosis of the initial pool.
  \emph{No kurtosis} uses
  $\alpha=0$, which reaches $\|\hat\kappa_4\|_F\le0.002$. \emph{Partial kurtosis}
  uses $\alpha=0.3$, giving $\|\hat\kappa_4\|_F=2.39$ against $7.95$ for its starting pool.
\end{itemize}
\section{Cumulants Review}
\label{app:cumulants}

Our results are phrased in terms of the cumulants of the task pool.
Here, we review cumulants and the few facts the main text uses.

\paragraph{Definition.}
Write $\mathbb{E}_{\taskspacetr}[g(\vbeta)] = \frac{1}{T}\sum_{t=1}^T g(\vbeta^{(t)})$ for the average over the uniform measure on the pool.
Its cumulant generating function is
\begin{equation}
    \hat K(\mathbf{t}) = \log \mathbb{E}_{\taskspacetr}\!\left[e^{\mathbf{t}^\top \vbeta}\right],
\end{equation}
and the order-$\ell$ cumulant $\hat\kappa_\ell$ is the symmetric tensor of its $\ell$-th derivatives at the origin,
\begin{equation}
    (\hat\kappa_\ell)_{a_1 \cdots a_\ell} = \frac{\partial^\ell \hat K}{\partial t_{a_1} \cdots \partial t_{a_\ell}}\bigg|_{\mathbf{t} = 0}.
\end{equation}
We write $M_\ell = \mathbb{E}_{\taskspacetr}[\vbeta^{\otimes \ell}]$ for the raw moment tensors. 

\paragraph{Relation to moments.}
With $\delta\vbeta = \vbeta - \hat\mu_\vbeta$ the centered task, the first four cumulants are
\begin{align}
    \hat\kappa_1 &= \hat\mu_\vbeta, \\
    \hat\kappa_2 &= \mathbb{E}_{\taskspacetr}\!\left[\delta\vbeta\,\delta\vbeta^\top\right] = \hat\Sigma_\vbeta, \\
    (\hat\kappa_3)_{abc} &= \mathbb{E}_{\taskspacetr}\!\left[\delta\beta_a\,\delta\beta_b\,\delta\beta_c\right], \\
    (\hat\kappa_4)_{abcd} &= \mathbb{E}_{\taskspacetr}\!\left[\delta\beta_a\,\delta\beta_b\,\delta\beta_c\,\delta\beta_d\right]
    - \big(\hat\Sigma_{ab}\hat\Sigma_{cd} + \hat\Sigma_{ac}\hat\Sigma_{bd} + \hat\Sigma_{ad}\hat\Sigma_{bc}\big).
\end{align}
In general, $\hat\kappa_\ell$ is the order-$\ell$ central moment with all relevant lower-order information subtracted---it is the information in the order-$\ell$ moment not already carried by lower orders.

\paragraph{Key properties used throughout our findings.}
Cumulants beyond the first are invariant under shifts, and all cumulants are multilinear under linear maps: if $\tilde\vbeta = A\vbeta$, then
\begin{equation}
    (\tilde\kappa_\ell)_{a_1 \cdots a_\ell} = \sum_{b_1, \ldots, b_\ell} A_{a_1 b_1} \cdots A_{a_\ell b_\ell}\, (\hat\kappa_\ell)_{b_1 \cdots b_\ell}.
    \label{eq:cum-multilin}
\end{equation}
Taking $A = \Sigma_\vx^{1/2}$ gives the colored cumulants $\tilde\kappa_\ell$ of \cref{app:correspondence}.

A Gaussian has all its cumulants beyond order two vanish.
The prior $\mathcal{N}(\mu_\vbeta, \Sigma_\vbeta)$ therefore carries only a mean and a covariance, but a \emph{finite} pool of $T$ draws from it does not.
Each $\hat\kappa_\ell$ is a sample statistic that converges to the prior's value as $T \to \infty$ with entrywise fluctuations of order $T^{-1/2}$, so $\hat\kappa_1 \to \mu_\vbeta$, $\hat\kappa_2 \to \Sigma_\vbeta$, and $\hat\kappa_{\ell \ge 3} \to 0$.
This rate is entrywise at fixed $d$: tensor norms carry additional factors of $d$, and when $d$ is comparable to $T$ the spectrum of $\hat\Sigma_\vbeta$ follows the Marchenko--Pastur law rather than concentrating at the prior's eigenvalues (\cref{subsec:taskdiv}).

Several eigencoefficients depend on cumulants only through partial traces.
We write $\mathrm{Tr}_k$ for the contraction of $k$ index pairs,
\begin{equation}
    (\mathrm{Tr}_k\, \hat\kappa_\ell)_{a_{2k+1} \cdots a_\ell} = \sum_{b_1, \ldots, b_k} (\hat\kappa_\ell)_{b_1 b_1 \cdots b_k b_k\, a_{2k+1} \cdots a_\ell},
    \label{eq:cum-trace}
\end{equation}
which by symmetry of $\hat\kappa_\ell$ does not depend on which pairs are contracted.
For example, $\mathrm{Tr}_1 \hat\kappa_3$, with components $\sum_a (\hat\kappa_3)_{aab}$, enters the degree-3 coefficients, and $\mathrm{Tr}_2 \hat\kappa_5$, with components $\sum_{a,b} (\hat\kappa_5)_{aabbc}$, enters the degree-5 coefficients.
We call a pool \emph{traceless} at order $\ell$ if $\mathrm{Tr}_1 \hat\kappa_\ell = 0$.
Contracted cumulants can therefore appear at lower polynomial degree than the full tensor, which first appears at degree $2\omega - 1$ for $\hat\kappa_\omega$ (\cref{thm:cum-aniso}).
\section{Review of the KRR Eigenframework for Predicting Test Performance from Task Eigenstructure}
\label{app:krr_eigenframework}

The central piece of existing theory which we use in this paper is a set of equations which give the average-case test MSE of kernel ridge regression (KRR) in terms of the task eigenstructure.
In this appendix, we will review this KRR eigenframework.

This eigenframework has been derived many times by different means in both the statistical physics community and the classical statistics community (which usually phrases the result as applying to linear ridge regression).
References studying KRR include \citet{sollich:2001-mismatched,bordelon:2020-learning-curves,jacot:2020-KARE,simon:2021-eigenlearning,loureiro:2021-learning-curves,wei:2022-more-than-a-toy}; references studying linear ridge regression include \citet{caponnetto:2007-decay-rates,dobriban:2018-ridge-regression-asymptotics,wu:2020-optimal,hastie:2022-surprises,richards:2021-asymptotics,cheng:2022-dimension-free-ridge-regression}.
The result is essentially the same in all cases.
Here we will adopt the terminology and notation of \citet{simon:2021-eigenlearning}.

Recall that we are studying KRR with a kernel function $K$ with data sampled $\vx_i \sim \mu$, targets generated as $y_i = f_*(\vx_i) + \eta$, and noise $\eta \sim \mathcal{N}(0, \epsilon^2)$ with variance $\epsilon^2 \ge 0$.
The kernel admits an eigendecomposition $K(\vx, \vx') = \sum_i \lambda_i \phi_i(\vx) \phi_i(\vx')$ with orthonormal eigenfunctions $(\phi_i)$.
Let us decompose the target function in the eigenbasis as $f_*(\vx) = \sum_i v_i \phi_i(\vx)$.
Suppose we run KRR with $n$ samples and a ridge parameter $\delta \ge 0$, and we compute the population (i.e. test) and train MSEs as
\begin{align}
    \texttt{MSE}_\textnormal{te, ID} &= \E{\vx \sim \mathcal{D}}{(f_*(\vx) - \hat{f}(\vx))^2} + \epsilon^2, \\
    \texttt{MSE}_\textnormal{tr} &= \frac{1}{n} \sum_i (y_i - \hat{f}(\vx_i))^2.
\end{align}

\subsection{Statement of the eigenframework}

We are now ready to state the eigenframework.
Let $\zeta \ge 0$ be the unique nonnegative solution to
\begin{equation}
    \label{eqn:kappa_def_eqn_app}
    \sum_i \frac{\lambda_i}{\lambda_i + \zeta} + \frac{\delta}{\zeta} = n.
\end{equation}
Then test risk is given approximately by
\begin{equation}
    \label{eqn:test_risk_eqn_app}
    \texttt{MSE}_\textnormal{te} \approx \mathcal{E}_\textnormal{te} := \mathcal{E}_0 \mathrm{B},
\end{equation}
where the \textit{overfitting coefficient} $\mathcal{E}_0$ is given by
\begin{equation}
    \label{eqn:overfitting_coeff_eqn_app}
    \mathcal{E}_0 := \frac{n}{n - \sum_i \left( \frac{\lambda_i}{\lambda_i + \zeta} \right)^2}
\end{equation}
and the \textit{bias} is given by
\begin{equation} \label{eqn:bias_eqn_app}
    \mathrm{B} = 
    \sum_i \left(
    \frac{\zeta}{\lambda_i + \zeta}
    \right)^2 v_i^2 + \epsilon^2.
\end{equation}
Train risk is given by
\begin{equation}
    \label{eqn:train_risk_eqn_app}
    \texttt{MSE}_\textnormal{tr} \approx \mathcal{E}_\textnormal{tr} \equiv \frac{\delta^2}{n^2 \zeta^2} \mathcal{E}_\textnormal{te},
\end{equation}

What is meant by the ``$\approx$'' in \cref{eqn:test_risk_eqn_app,eqn:train_risk_eqn_app}?
This result only becomes exact in certain stringent cases; it is formally derived under an assumption that the eigenfunctions are independent Gaussian (or sub-Gaussian) variables when $\vx$ is sampled from $\mu$, and it is exact only in an asymptotic limit in which $n$ and the number of eigenmodes in a given eigenvalue range (or the number of duplicate copies of any given eigenmode) both grow large at a proportional rate \citep{hastie:2022-surprises,bach:2023-rf-eigenframework}.
These conditions emphatically do not apply to any realistic instance of KRR.
Nonetheless, numerical experiments find that \cref{eqn:test_risk_eqn_app,eqn:train_risk_eqn_app} hold with small error even at modest $n$ \citep{canatar:2021-spectral-bias,simon:2021-eigenlearning,wei:2022-more-than-a-toy}: though derived in an idealized setting and exact only in a limit, this eigenframework holds reliably in practical cases.

\begin{figure}[ht!]
    \includegraphics[width=\textwidth]{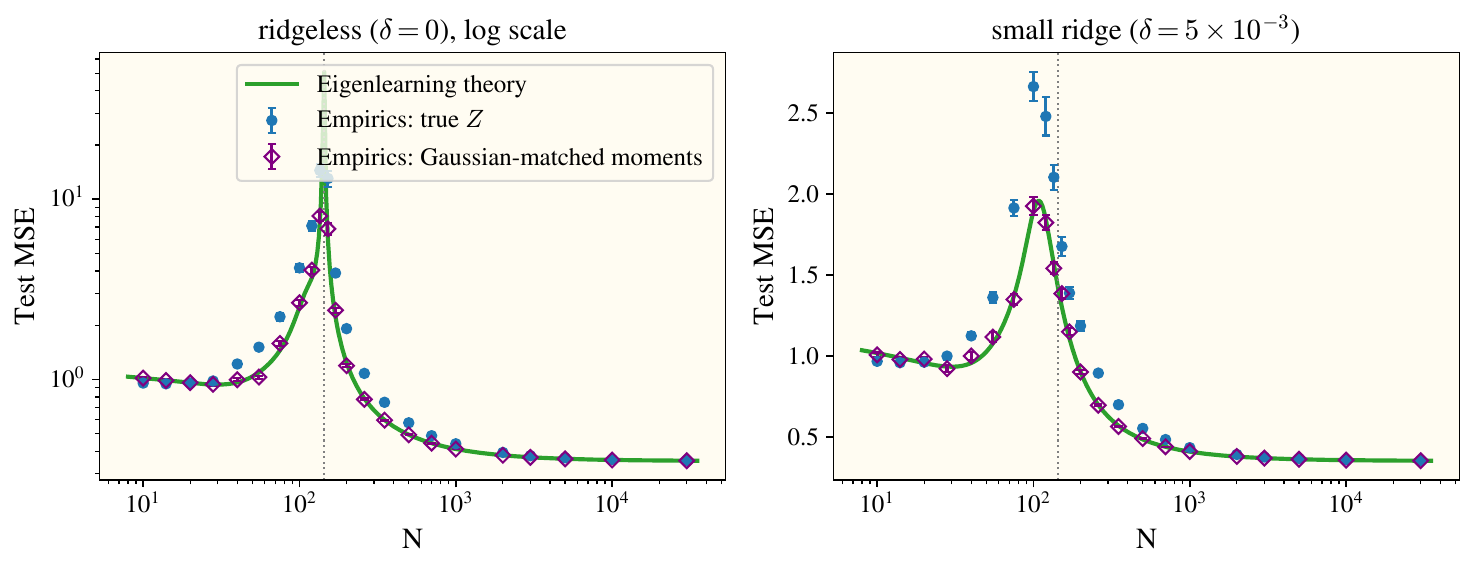}
        \caption{
        \textbf{One-dimensional ICL data breaks Gaussian universality.}
        We compare the eigenframework's predicted test MSE (line) against kernel regression trained on true ICL contexts (circles) and on a Gaussian surrogate with matched moments (diamonds), for $T=4$, $d=1$, $L=8$.
        We use a Gaussian kernel with kernel width $\sigma_\text{ker} = \sqrt3$ and in-context data with covariance $\Sigma_\vx \sim I_d/(dL)$.
        The left panel is ridgeless ($\delta=0$) and the right uses a small ridge ($\delta=5\times10^{-3}$).
        The Gaussian surrogate tracks the theory at every $n$.
        The true contexts agree away from the interpolation peak but exceed the prediction near it, most visibly at small ridge.
        Both datasets share the same theoretical curve, so the mismatch comes from the non-Gaussianity of one-dimensional ICL data rather than from the predicted eigenstructure. We observe no such failure for $d>1$.
        \label{appfig:gaussian_universality}
        }
\end{figure}

In this paper, we use this eigenframework as a tool to map predictions of task eigenstructure to predictions of learning curves.
Since we are here using it in settings very similar to those tested by previous works \citep{bordelon:2020-learning-curves,jacot:2020-KARE,simon:2021-eigenlearning,wei:2022-more-than-a-toy}, we expect it to work well.
Its use introduces some small error (as it is not perfect at finite $n$), but this is usually not the dominant source of error.
We have found numerically that this framework fails on ICL problems with a single intrinsic dimension, $d=1$, but works well otherwise.
This failure mode is due to the Gaussian universality condition breaking; see \cref{appfig:gaussian_universality}.

\subsection{The Out-of-Distribution Eigenframework}

While the eigenframework as presented in \citet{sollich:2001-mismatched,bordelon:2020-learning-curves,jacot:2020-KARE,simon:2021-eigenlearning,loureiro:2021-learning-curves,wei:2022-more-than-a-toy} specify only the learning dynamics when the distribution of the trainset is equivalent to that of the testset, our techniques require specifying the test error on a novel ``ICL'' distribution.
Luckily, an out-of-distribution version of the eigenframework appears in \citet{canatar2022outofdistributiongeneralizationkernelregression}, which we can use as a baseline, choosing to adapt the same notation as \citet{simon:2021-eigenlearning}.
The main difference between the in-distribution and an out-of-distribution test set is that our eigenfunctions need no longer be \textit{eigen} (orthonormal) on our new distribution.
Therefore, the bulk of our work will be in describing this non-orthonormality.
We begin by defining two useful quantities going forward, the eigenfunction overlap and the test eigencoefficients:

\begin{align}
    S_{ij} &= \left< \phi_i(Z)\phi_j(Z)\right>_{\mathcal{D}_\text{OOD}}\\
    \tilde{v}_i &= S_{ij}^{-1}\left< y^\star \phi_j(Z)\right>_{\mathcal{D}_\text{OOD}}
\end{align}

Writing $y^\star$ as a sum of a resolvable and unresolvable portion,

\begin{equation}
    y^\star(Z) = \sum_i\tilde{v}_i\phi_i(Z) + f_\perp,
\end{equation}

we can expand $\texttt{MSE}_\textnormal{te, OOD}$ as,

\begin{align}
    \texttt{MSE}_\textnormal{te, OOD} &= \E{\vx \sim \mathcal{D}_\text{OOD}}{(f_*(\vx) - \hat{f}(\vx))^2} + \epsilon^2\\
    &= \E{\vx \sim \mathcal{D}_\text{OOD}}{(\hat{v}-\tilde{v})^\top S(\hat{v}-\tilde{v})} + ||f_\perp||^2_\text{OOD}+\epsilon^2\\
    &=(\mathbb{E}[\hat{v}-\tilde{v}])^\top S (\mathbb{E}[\hat{v}-\tilde{v}]) + \text{tr}[ S \cdot Cov(\hat{v})] + ||f_\perp||^2_\text{OOD}+\epsilon^2\\
    &=\Delta^\top S \Delta + \frac{\mathcal{E}_0}{n}\sum_i S_{ii}\mathcal{L}_i^2 \left(\sum_j (1-\mathcal{L}_j)^2v^2_j+\epsilon^2\right)+||f_\perp||^2_\text{OOD}+\epsilon^2
\end{align}

where $\Delta = \tilde{v}-\mathcal{L}v$.
The first part of the equation ($\Delta^\top S \Delta$) can be understood to be the ``out-of-distribution bias'' term, the second part as the ``out-of-distribution variance'' term, and the last ($||f_\perp||^2_\text{OOD}+\epsilon^2$) as an unresolvable noise.
\section{Details on the Kernel Eigenstructure}
\label{app:eigensystem}

\begin{figure}[t]
    \includegraphics[width=\textwidth]{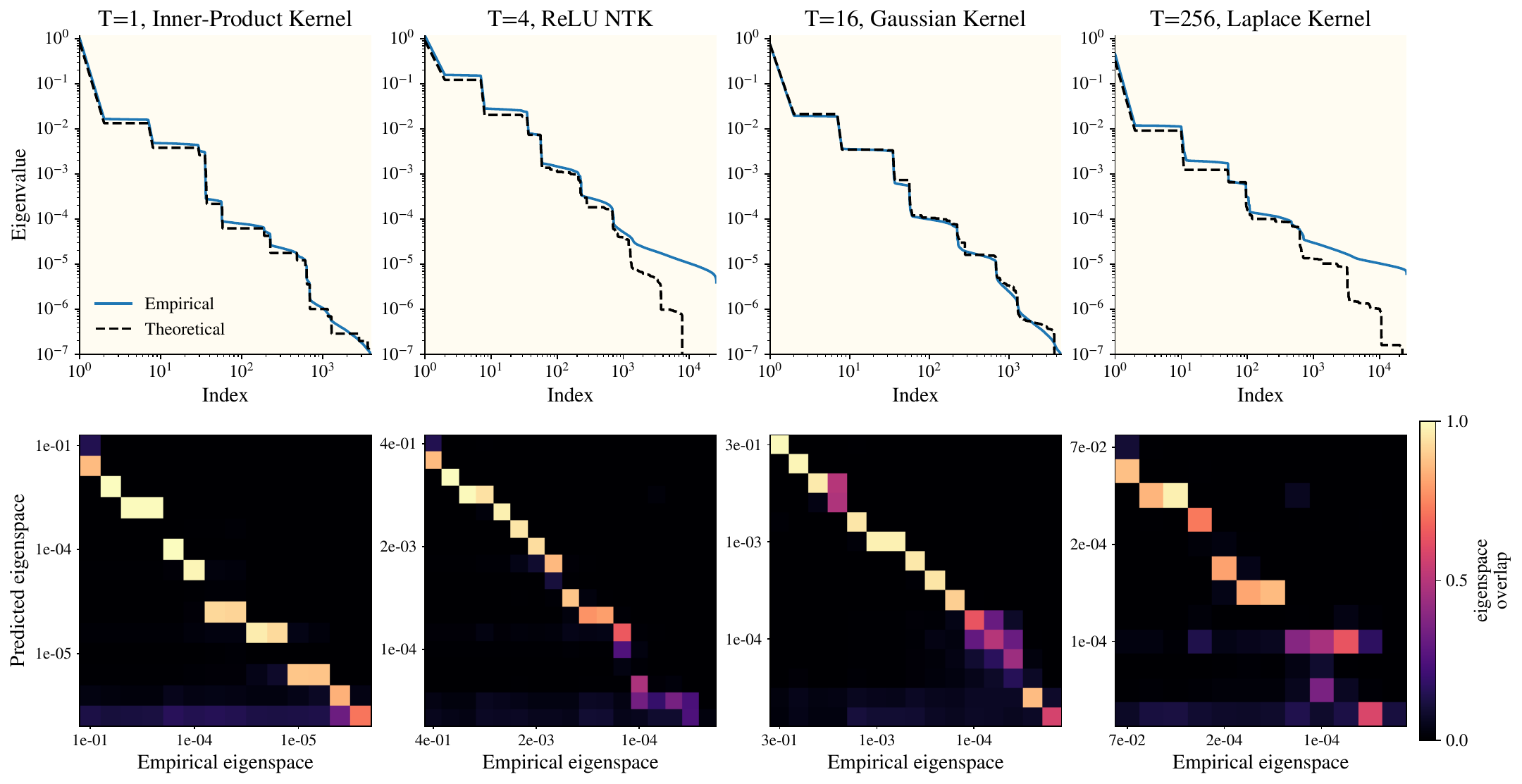}
    \caption{
    \textbf{The subkernel expansion framework (\cref{def:see}) accurately predicts the eigenstructure of in-context learning data for various rotation-invariant kernels.}
    Across various rotation-invariant kernels with varying task diversity (columns), we compute the empirical kernel eigensystem $\{(\lambda^{(\text{emp})}_i, \phi^{(\text{emp})}_i)\}$ and compare to the theoretical eigensystem $\{(\lambda^{(\text{th})}_i, \phi^{(\text{th})}_i)\}$ obtained from \cref{def:see}, indexed in order of decreasing $\lambda^{(\text{th})}_i$.
    In the top row, we overlay the eigenvalue spectrum between $\lambda^{\text{(emp})}_i$ and $\lambda^{\text{(th})}_i$.
    In the bottom row, we bin both the predicted and empirical eigenfunctions into shared spectral bins and visualize the pairwise subspace overlap (\cref{eq:overlap}).
    Concentration along the diagonal indicates theory-experiment match.
    See \cref{app:experiment_details} for further explanation and experimental details.
    }
    \label{fig:eigendecom}
\end{figure}

The kernel's prediction is governed by its (eigenvalue, eigenfunction) pairs $(\lambda_i, \phi_i)$ with respect to the training distribution $\mathcal{D}_\mathrm{tr}$,
\begin{equation}
    \mathbb{E}_{\vz' \sim \mathcal{D}} \left[K(\vz, \vz') \phi_i(\vz')\right] = \lambda_i \phi_i(\vz),
    \qquad
    \mathbb{E}_{\vz \sim \mathcal{D}} \left[\phi_i \phi_j\right] = \delta_{ij}.
    \label{eq:ker_eigsys}
\end{equation}

This is analytically intractable for the ICL data and kernels that we analyze, so we need instead to find the eigendecomposition another way.
Here, we walk through how to arrive at the subkernel expansion framework, our method of approximating the eigensystem, then give its full definition.
The ICL regression task does not have closed-form eigenfunctions in full generality, but we can nonetheless approximate them well.

\subsection{The subkernel expansion framework}
\label{appsubsec:sexpansion}

We first give a brief primer on what the subkernel expansion framework is at a high level before giving the technical definition, much in the same way integrals are first introduced using Riemann sums.

Because $\|\vz\|$ concentrates for ICL contexts, the rotation-invariant kernels are well approximated by inner-product kernels (\cref{def:inner-product-kernel}), which depend on their inputs only through $\vz^\top \vz'$.
% (\cref{app:eigensystem})
\begin{definition}[Inner-product kernel and its subkernels]
    A kernel is an \emph{inner-product kernel} if $K(\vz, \vz') = \sum_{\ell \ge 0} \frac{c_\ell}{\ell!} (\vz^\top \vz')^\ell$ with ``level coefficients'' $c_\ell \ge 0$, which guarantees $K$ is PSD.
    Its degree-$\ell$ \emph{subkernel} is the single term $K_\ell(\vz, \vz') = \frac{c_\ell}{\ell!}(\vz^\top \vz')^\ell$.
    \label{def:inner-product-kernel}
\end{definition}

We begin with an intuitive explanation for the eigenfunctions.
We treat each subkernel of \cref{def:inner-product-kernel} separately.
The degree-1 subkernel is nothing but a linear kernel (\textit{i.e.} performing linear regression), so its eigenfunctions are the principal components of the raw data.
Since $(\vz^\top\vz')^\ell = (\vz)_\ell^\top (\vz')_\ell$, where $(\vz)_\ell$ collects all degree-$\ell$ polynomials of $\vz$, the degree-$\ell$ subkernel is likewise a linear kernel on the degree-$\ell$ polynomials of the data, and we might expect its eigenfunctions to be their principal components.
This fails because polynomials overlap \textit{across} degrees (\textit{e.g.} $x^2$ with the constant).
To handle this, we impose the assumption that level coefficients decay quickly, $c_{\ell+1}\,\mathbb{E}\|\vz\|^2 \ll c_\ell$.
Each shared component is then claimed by the lowest degree that contains it.
The degree-$\ell$ eigenfunctions are therefore the principal components of the degree-$\ell$ polynomials after all lower degrees are projected out, with eigenvalues as $c_\ell/\ell!$ times the variances of those components.

\paragraph{The terms of the subkernel expansion with ICL data}

In the walkthrough, we labeled the terms simply as ``polynomials,'' although it's helpful to discuss which polynomials we are talking about, and which will be used for a model's computation.
We find that only a select subset will have non-0 coefficient when expanding functions in the kernel's eigenbasis, which we label as subkernel expansion ICL polynomials.
These serve as good approximations to the eigenfunctions, with the upside that they're still interpretable.

\begin{definition}[Subkernel expansion ICL polynomials]
    The degree $p$ Subkernel expansion ICL polynomials, $\Psi_{p}(\vz)$, are a set of functions $\psi_p$ that can be written as a tensor product of $p$ components of the context $\vz$ in the following manner:
    \begin{equation}
        \psi_{p}(\vz) \equiv \left( \bigotimes_{i=1}^m \sum_{\ell=1}^L y_\ell^{a_i} \vx_\ell^{\otimes b_i} \right) \otimes \vx_{L+1}
    \end{equation}
    where $p= 1+\sum_{i=1}^{m}\left(a_i+b_i\right)$.
    $\Psi_p$ contains multiple different polynomial functions, which we enumerate with $\psi_p, \psi_p', \psi_p^{(2)},\dots, \psi_p^{(k)}$.
    % $\omega=1+\sum_i a_i$, $k$ is the number of self-contracted indices, ones where $a_i\ge 2, b_i=0$, and $m$ is the number of independent context factors.
    For instance, the degree $3$ Subkernel expansion ICL polynomials contain both the functions $\psi = \sum_i y_i\vx_i\otimes\vx_{L+1}$ of size $d^2$ and $\psi_3' = \sum_i y_i^2 \vx_{L+1}$ of size $d$.
    % ($2$, $0$) family consists of $d^2$ functions specified by $\big\{\sum_i y_i x_{i,\alpha}\,x_{L+1,\gamma}\big\}_{\alpha,\gamma=1}^{d}$.
    % We occasionally specify the Subkernel expansion ICL polynomials by their total degree $p$: $m= 1+\sum_{i=1}^{m}\big(a_i+b_i\big)$.
    \label{def:regpoly}
\end{definition}
Although it would be mathematically simple to work with Subkernel expansion ICL polynomials, we cannot, as they couple across different orders.
For instance, $\Psi_{5}$ overlaps nontrivially with $\Psi_{3}$, e.g. $\psi_5 = \sum_i y_i^3 \vx_i\vx_{L+1}$ and $\psi_3 = \sum_i y_i \vx_i\vx_{L+1}$ are not orthogonal as $\mathbb{E}_\mathcal{\d}[y_i^2] > 0$.
To account for this overlap, we invoke that our kernels will learn the optimal low-degree function first (\textit{i.e.} the lower-degree functions will have larger eigenvalues), and therefore project out the lower-degree information from the higher degree polynomial.
We label these processed terms as the ``Subkernel expansion ICL perpendicular polynomials.''
\begin{definition}[Subkernel expansion ICL perpendicular polynomials]
    A degree $p$ set of subkernel expansion ICL perpendicular polynomials, $\Psi_{p}^\perp(\vz)$, is constructed as a corresponding subkernel expansion ICL polynomial (\cref{def:regpoly}) with all overlap between lower-degree subkernel expansion ICL polynomials projected out:
    \begin{equation}
        \Psi^\perp_p(\vz) \equiv \Psi_p(\vz) - \mathbb{E}_{\mathcal{D}}\left[\Psi_p \Psi_{<p}^\top\right]\mathbb{E}_{\mathcal{D}}\left[\Psi_{<p}\Psi_{<p}^\top\right]^{+}\Psi_{<p}(\vz)
        % \Psi_{p}^\perp(Z) \equiv \Psi_{p}(Z) - \sum_{\{p'\} < p}\mathbb{E}_Z[\Psi_{p}(Z)\Psi^\top_{\{p'\}}(Z)]{\mathbb{E}_Z\left[\Psi_{\{p'\}}(Z)\Psi_{\{p'\}}^\top\right]^{-1}}\Psi_{\{p'\}}(Z),
    \end{equation}
    where each function of $\Psi_p$ is subtracted against \emph{every} subkernel expansion ICL polynomial of a lower degree, and $(\cdot)^+$ denotes a pseudoinverse.
    To note, we do not make any subtractions within a degree.
    % We make the assumption that the different subsets ${\psi^{(k)}}^\perp$ for any $(k)$ are orthogonal, which holds exactly as $L\to\infty$.
    % where $\{p'\}$ is used to denote every Subkernel expansion ICL polynomial family with degree $p'$.
    \label{def:perppoly}
\end{definition}
The subkernel expansion ICL perpendicular polynomials are now orthogonal across degrees, but not within a given degree: %
the polynomials share within-context fluctuations and become correlated through the task distribution's statistics. 
% The Subkernel expansion ICL perpendicular polynomials are still not orthonormal.
%at least not within their own degree. 
% 
% Orthonormal functions $\phi$ need to satisfy $\mathbb{E}_Z[\phi_i \phi_j]=\delta_{ij}$, which we don't have for the Subkernel expansion ICL perpendicular polynomials: %
For example, the overlap between $\psi_3^\perp = \sum_i y_i\vx_{i,1} \vx_{L+1, 1}\approx \sum_i \vbeta_1^\top I_d\vx_{L+1,1}/d$ and $\psi_3^\perp = \sum_i y_i\vx_{i,2} \vx_{L+1, 1}\approx \sum_i \vbeta_2^\top I_d\vx_{L+1,1}/d$ is non-zero unless the covariance of the training tasks is isotropic ($\hat\Sigma_\vbeta \propto I_d$).
To account for this, we orthonormalize each degree as a single block using its Gram matrix to obtain the kernel's predominately useful eigenfunctions, the ones with non-0 eigencoefficient.

\paragraph{The subkernel expansion framework, in full detail}

Here, we provide the subkernel expansion framework.

\boxit{
\begin{definition}[Subkernel expansion]
    The subkernel expansion of a kernel $K$ is obtained using the following procedure:
    \begin{enumerate}
    \item Expand the chosen kernel function as an inner-product kernel (\cref{def:inner-product-kernel}) with level coefficients $c_\ell$. 
    \item Let $(\vz)_\ell$ be the vector of all entries of $\vz^{\otimes\ell}$, i.e.\ all ordered products of $\ell$ coordinates of $\vz$, so that $(\vz^\top\vz')^\ell=(\vz)_\ell^\top(\vz')_\ell$ exactly.
    Let $(\vz)_{<\ell}$ stack $(\vz)_0,\dots,(\vz)_{\ell-1}$.
    For convenience, denote all polynomials at the current level $\ell$ with the marker ``F,'' and polynomials at all levels below the current level with the marker ``L.''
    \item Construct the second-moment matrix of the degree-$\ell$ features, $C_{FF} \equiv \mathbb E\bigl[(\vz)_\ell(\vz)_\ell^\top\bigr]$.
    \item Similarly, construct $C_{FL} \equiv \mathbb E\bigl[(\vz)_\ell(\vz)_{<\ell}^\top\bigr]$ between the degree-$\ell$ features and all lower-degree ones, and $C_{LL} \equiv \mathbb E\bigl[(\vz)_{<\ell}(\vz)_{<\ell}^\top\bigr]$.
    \item Project out all lower-degree features using the Schur complement, $\tilde{C}_\ell = C_{FF}-C_{FL}C_{LL}^{+}C_{FL}^\top$, where $^+$ is the pseudoinverse ($C_{LL}$ is singular because reordered products repeat).
    \item The subkernel $K_\ell$'s eigendecomposition is equivalent to a diagonalization of $\frac{c_\ell}{\ell!}\tilde{C}_\ell$.
    \item The eigendecomposition of the original kernel $K$ is the set of eigendecompositions of each subkernel:
    \begin{equation*}
        \text{Eigendecomp}(K) \approx \text{Eigendecomp}(K_0)+\text{Eigendecomp}(K_1)+\dots\,
    \end{equation*}
    \end{enumerate}
    \label{def:see}
\end{definition}
}

\paragraph{Note on Hermite polynomials}
For Gaussian data, the level-$\ell$ eigenfunctions are Hermite polynomials of degree $\ell$ in the data's principal components \citep{karkada2026predictingkernelregressionlearning}.
ICL data is exactly Gaussian only for a single task ($T=1$), where each label is a linear function of its input plus Gaussian noise; there, the subkernel expansion reduces to their Hermite structure.
For $T>1$, mixing over tasks makes the context non-Gaussian and correlates distinct positions through the shared task (the matrix $C$ of \cref{apppara:quadratic}), so the eigenfunctions depart from Hermite polynomials.
The exception is the query, which is Gaussian and independent of the context, so its factor remains Hermite (\cref{prop:M}).

\subsection{Potential proof sketch for the subkernel expansion framework}
\label{appsubsec:sex_proof}

We have not proven the subkernel expansion framework in general; it is validated experimentally in \cref{fig:eigendecom}.
Nonetheless, we sketch a possible proof below.

\paragraph{Proof sketch main idea.}
Let $\mu$ be any data measure with $\|\vz\|^2\le R^2$, and let $K(\vz,\vz')=\sum_\ell \frac{c_\ell}{\ell!}(\vz^\top \vz')^\ell$ with $c_{\ell+1}R^2\le \varepsilon c_\ell$.
For every fixed level $\ell$, the true kernel eigenvalues match the level-$\ell$ subkernel expansion eigenvalues (with the same indices) up to an additive $O(\varepsilon c_\ell R^{2\ell}/\ell!)$. This is a relative $1+O(\varepsilon)$ for eigenvalues of that order.
The true eigenspaces match the framework's eigenspaces up to an overlap error of $O(\varepsilon)$ divided by the relative spectral gap.
For the Gaussian kernel, $c_{\ell+1}/c_\ell=\sigma^{-2}$, so $\varepsilon\approx R^2/\sigma^2$.

The argument bounds each eigenvalue from above and below; the eigenspace overlap then follows by standard perturbation theory.

\emph{Upper bound.}
Apply Courant--Fischer on the functions orthogonal to $\mathcal{P}_{<\ell}$ and to the top $j-1$ level-$\ell$ framework eigenfunctions.
There, the lower subkernels vanish, level $\ell$ contributes exactly the framework's quadratic form, and the higher levels contribute at most $\sum_{m>\ell}c_mR^{2m}/m! = O(\varepsilon c_\ell R^{2\ell}/\ell!)$.

\emph{Lower bound.}
Assume the lower-level eigenvalues are of order $c_{\ell-1}R^{2(\ell-1)}/(\ell-1)!$, which by hypothesis is at least of order $c_\ell R^{2\ell}/(\varepsilon\,\ell!)$.
The coupling between $\mathcal{P}_{<\ell}$ and the level-$\ell$ functions comes through $K_\ell$ and the higher levels, and is of order $c_\ell R^{2\ell}/\ell!$.
Eliminating the lower block therefore shifts the eigenvalue by at most $O(\varepsilon c_\ell R^{2\ell}/\ell!)$.

\subsection{Using the subkernel expansion framework with ICL data}

Above, we introduced the subkernel expansion.
Now, we apply it to the ICL linear regression setting to find the eigensystem beyond heuristics.

We begin with $d=1$ data for readability, so each $x_i$ is a scalar with variance $\sigma_x^2$, and we write $m_k=\frac1T\sum_{t=1}^T\beta_t^k$ for the raw moments of the pool.

\paragraph{Linear subkernel.}
The level-$1$ subkernel $K_1=c_1\vz^\top\vz'$ performs linear regression. Its eigenfunctions are the principal components of the data's second-moment matrix, with eigenvalues $c_1$ times those of the matrix.
Every coordinate of $\vz$ has mean zero, so no projection against the constant is needed, and the second-moment matrix is the covariance.
Distinct positions are uncorrelated: $\mathrm{Cov}(x_i,x_j)=\mathrm{Cov}(x_i,y_j)=\mathrm{Cov}(y_i,y_j)=0$ for $i\neq j$, and the query is uncorrelated with every in-context coordinate.
The covariance is therefore block diagonal.
It consists of an isolated query block, with eigenfunction $x_{L+1}$ and eigenvalue $c_1\sigma_x^2$, and $L$ identical $2\times2$ blocks, one per position:
\begin{equation*}
\mathrm{Cov}\begin{pmatrix} x_i\\ y_i\end{pmatrix}=
\begin{pmatrix}
    \mathbb{E}_{\mathcal D}[x_i^2] & \mathbb{E}_{\mathcal D}[x_i(\beta x_i+\eta_i)]\\
    \mathbb{E}_{\mathcal D}[x_i(\beta x_i+\eta_i)] & \mathbb{E}_{\mathcal D}[(\beta x_i+\eta_i)^2]
\end{pmatrix}
=
\begin{pmatrix}
    \sigma_x^2 & m_1\sigma_x^2\\
    m_1\sigma_x^2 & m_2\sigma_x^2+\epsilon^2
\end{pmatrix},
\end{equation*}
where the expectation over $\mathcal D$ averages $\beta$ over the finite pool.
Each block has eigenvalues and (unnormalized) eigenfunctions
\begin{align*}
    \lambda_\pm &= \frac{1}{2}\left(\sigma_x^2(1+m_2)+\epsilon^2\pm\sqrt{\bigl(\sigma_x^2(1-m_2)-\epsilon^2\bigr)^2+4m_1^2\sigma_x^4}\right),\\
    \phi_{\pm,i} &\propto m_1\sigma_x^2\,x_i+(\lambda_\pm-\sigma_x^2)\,y_i,
\end{align*}
so the subkernel eigenvalues $c_1\lambda_\pm$ are each $L$-fold degenerate.
Of these linear eigenfunctions, only $x_{L+1}$ carries a nonzero target coefficient (the mean $\tilde\mu$ of \cref{apptable:eigcoeffs}).

\paragraph{Quadratic subkernel.}
\label{apppara:quadratic}
Degree-2 polynomials have nonzero means, so at level 2 we must project out the constant; in practice this means centering each monomial.
Projection against level 1 is automatic, since every third moment vanishes under the symmetry $\vz\mapsto-\vz$.
The multiplicities matter here.
Since $(\vz^\top\vz')^2=\sum_{\mu\nu}z_\mu z_\nu z'_\mu z'_\nu$, a product of two distinct coordinates appears twice, so its feature carries a factor $\sqrt2$.
With that convention, the eigenvalues of $K_2$ are $c_2/2!$ times those of the feature covariance.
The covariance splits into blocks:
\begin{enumerate}
    \item $x_{L+1}^2-\sigma_x^2$ with itself.
    \item $\sqrt2\,x_{L+1}x_i$ with $\sqrt2\,x_{L+1}y_i$: the linear-subkernel block with a factor of $x_{L+1}$ attached, giving eigenvalues $2\sigma_x^2\lambda_\pm$, each $L$-fold degenerate.
    \item For each pair $i<j$, the $4\times4$ block of $\sqrt2\,x_ix_j$, $\sqrt2\,x_iy_j$, $\sqrt2\,y_ix_j$ and $\sqrt2\,y_iy_j$, of which there are $L(L-1)/2$ identical copies.
    Because both positions share the same task, this block is $2\,\mathbb E_\beta[B(\beta)\otimes B(\beta)]$, where $B(\beta)$ is the conditional covariance of $(x_i,y_i)$ given $\beta$; it is not the Kronecker square of the pool-averaged linear block.
\end{enumerate}

The remaining polynomials require more care.
Define the centered single-position features
\begin{equation*}
a_i = x_i^2-\sigma_x^2, \qquad
b_i = \sqrt2\,\bigl(x_iy_i-m_1\sigma_x^2\bigr), \qquad
c_i = y_i^2-\bigl(m_2\sigma_x^2+\epsilon^2\bigr).
\end{equation*}
Let $D$ be their $3\times3$ covariance at a single position, and $C$ their cross-covariance between two distinct positions $i\neq j$.
Neither depends on which positions are chosen.
Given $\beta$, distinct positions are independent, so $C$ is the pool covariance of the features' conditional means,
\begin{equation*}
C=\mathrm{Cov}_{\beta}\bigl(\mathbb E[(a_i,b_i,c_i)\,|\,\beta]\bigr)
=\sigma_x^4\begin{pmatrix}
    0 & 0 & 0\\
    0 & 2\,\mathrm{Var}\,\beta & \sqrt2\,\mathrm{Cov}(\beta,\beta^2)\\
    0 & \sqrt2\,\mathrm{Cov}(\beta,\beta^2) & \mathrm{Var}\,\beta^2
\end{pmatrix},
\end{equation*}
with variances and covariances taken over the pool.
In particular, $\mathbb E[b_ib_j]=2\sigma_x^4(m_2-m_1^2)$ vanishes only when every task in the pool is identical.
Unlike the linear case, distinct positions are therefore correlated through the shared task, and the features span a single $3L\times3L$ block, $I_L\otimes(D-C)+\mathbf 1\mathbf 1^\top\otimes C$.
Its eigensystem is the union of the following:
\begin{enumerate}
    \item $D+(L-1)C$, for functions summed over the context, $\sum_i(\cdot)_i$;
    \item $D-C$, which is $(L-1)$-fold degenerate, for contrasts between positions, such as $b_i-b_j$.
\end{enumerate}
The first sector contains the two eigenvalues that grow with $L$; its eigenfunctions are spanned by $\sum_i b_i$ and $\sum_i c_i$ and diagonalize the pool covariance of $(\beta,\beta^2)$.
The sum $\sum_i a_i$ does not grow, since the first row of $C$ vanishes.
These are the $d=1$ versions of the families $S$, $A$ and $W$ of \cref{appsubsec:whitened}, with $\beta$ and $\beta^2$ as the task objects of $S$ and $A$.

\paragraph{Cubic and higher subkernels.}
We do not track the higher subkernels in detail, but one property is key.
The ICL-relevant function $\sum_i x_iy_i\,x_{L+1}$ lies in the context-sum sector, whose eigenvalue grows linearly in $L$ through $C_{bb}=2\sigma_x^4\,\mathrm{Var}\,\beta$, and therefore needs only to be orthonormalized using the subkernel expansion framework.

\paragraph{In higher dimensional data}

The same procedure outlined above applies in $d$ dimensions.
Within a position, the label $y_i=\vbeta^\top\vx_i+\eta_i$ now couples to every coordinate of $\vx_i$, so the per-position blocks become $(d+1)\times(d+1)$.
Across positions, the structure is as in one dimension: distinct positions are uncorrelated at level 1, and at level 2 and above they are correlated only through the shared task.
\section{Eigencoefficients of the ICL Linear Regression problem}
\label{app:correspondence}

\subsection{Kernels are approximations of memorizing solutions}

In the main text we used the fact that a kernel learner's eigencoefficients are those of the memorizing solution.
This holds in any supervised setting, not only ICL: it is the standard statement that kernel regression targets the conditional mean under the training distribution.
We state it in the form we use.

\begin{lemma}[Kernel eigencoefficients are those of the memorizing solution]
    Let $\{\phi_i(\vz)\}$ be the ($\vz$-measurable) eigenbasis of a PSD kernel $K$.
    As $f_{\mathrm{mem}}(\vz) = \mathbb{E}[y| \vz]$ is the projection of $y^\star$ onto $\vz$-measurable functions, the eigencoefficients of the target and a memorizing predictor are mode-identical:
    \begin{equation}
      v_i \equiv \mathbb{E}_{(\vz, y)\sim \mathcal{D}_\mathrm{tr}}[f^\star(\vz)\phi_i(\vz)]= \mathbb{E}_{(\vz, y)\sim \mathcal{D}_\mathrm{tr}}[y\phi_i(\vz)]
      = \mathbb{E}_{(\vz, y)\sim \mathcal{D}_\mathrm{tr}}[f_{\text{mem}}(\vz)\phi_i(\vz)]
    \end{equation}
    for all $i$.
    Therefore, an infinite-sample kernel will implement the memorizing solution exactly, with a finite-sample kernel implementing an approximation based on the kernel's largest eigenvalues.
    \label{lem:kernel-memorize}
\end{lemma}

\begin{proof}
    Let $\mathcal{D}_\mathrm{tr}$ be the training measure on contexts, let $y^\star \in L^2$ be a target whose randomness is not exhausted by $\vz$, and let $f_\text{mem}(\vz) \equiv \mathbb{E}[y^\star | \vz]$ be the Bayes-optimal predictor under $\mathcal{D}_\mathrm{tr}$. For any orthonormal set $\{\phi_i\}$, the law of total expectation yields
    \begin{equation*}
    \mathbb{E}_{(\vz, y)\sim \mathcal{D}_\mathrm{tr}}[y^\star, \phi_i]
    = \mathbb{E}\left[\mathbb{E}\left[y^\star \phi_i(\vz) | \vz\right]\right]
    = \mathbb{E}\left[\phi_i(\vz)\mathbb{E}\left[y^\star | \vz\right]\right]
    = \mathbb{E}_{(\vz, y)\sim \mathcal{D}_\mathrm{tr}}[f_\text{mem}, \phi_i].
    \end{equation*}
    \label{proof:kernel-memorizing}
\end{proof}

\subsection{Whitened eigensystem}
\label{appsubsec:whitened}

The main text takes $\Sigma_\vx\propto I_d$.
General $\Sigma_\vx$ reduces to this case through the rescaling $\vbeta^\top\vx=(\Sigma_\vx^{1/2}\vbeta)^\top(\Sigma_\vx^{-1/2}\vx)$.
This invertible linear map leaves the space of polynomials of each degree unchanged, so the eigencoefficients are those of the isotropic case with the tasks replaced by the colored tasks $\Sigma_\vx^{1/2}\vbeta$, and the cumulants that enter are those of the colored pool.
The kernel itself is not whitened: its eigenvalues are still computed in raw coordinates and depend on $\Sigma_\vx$.

We denote whitened and colored quantities with a tilde,
\begin{equation}
    \tilde{\vx} = \Sigma_\vx^{-1/2}\vx, \qquad
    \tilde{\vbeta} = \Sigma_\vx^{1/2}\vbeta, \qquad
    \tilde{\mathcal{B}}_{\mathrm{tr}} = \mathcal{U}\left(\{\Sigma_\vx^{1/2}\vbeta^{(t)}\}_{t=1}^{T}\right),
\end{equation}
and write $\tilde\kappa_\ell$ for the cumulants of $\tilde{\mathcal B}_{\mathrm{tr}}$; by multilinearity (\cref{eq:cum-multilin}), $\tilde\kappa_\ell$ is $\hat\kappa_\ell$ with every index contracted against $\Sigma_\vx^{1/2}$.
We write $\tilde\mu$, $\tilde M_2$ and $\tilde\Sigma$ for the mean, second moment, and covariance of $\tilde{\mathcal B}_{\mathrm{tr}}$, $(\tilde s_a,\vu_a)$ for the eigenpairs of $\tilde\Sigma$, and $\Tr_k$ for the partial trace of \cref{eq:cum-trace}.
Whitening gives $\tilde{\vx}_i\sim\mathcal N(0,I_d)$, and the labels $y_i=\tilde\vbeta^\top\tilde\vx_i+\eta_i$ are unchanged.
We define the per-task label variance and its pool average,
\begin{equation}
    \tau(\vbeta) = \vbeta^\top\Sigma_\vx\vbeta + \epsilon^2 = \|\tilde{\vbeta}\|^2 + \epsilon^2, \qquad
    \bar\tau = \mathbb E_{\mathcal B_{\mathrm{tr}}}[\tau(\vbeta)] = \Tr(\tilde M_2) + \epsilon^2 .
\end{equation}

We collect the $L$ in-context pairs, but not the query, into the context matrix $\vz_{\mathrm c}=[\tilde X\;\boldsymbol y]\in\mathbb R^{L\times(d+1)}$, whose $i$-th row is $(\tilde\vx_i^\top, y_i)$, and form its Gram matrix
\begin{equation}
    M \equiv \vz_{\mathrm c}^\top \vz_{\mathrm c}=\begin{pmatrix} W & S\\ S^\top & A\end{pmatrix},\qquad
    W=\sum_{i=1}^L\tilde{\vx}_i\tilde{\vx}_i^\top,\quad S=\sum_{i=1}^L y_i\tilde{\vx}_i,\quad A=\sum_{i=1}^L y_i^2.
\end{equation}

\begin{proposition}[The target lives on polynomials in $M$]
    Given $\vbeta$, the rows of $\vz_{\mathrm c}$ are i.i.d.\ $\mathcal N(0,\Sigma_{\vz\mid\vbeta})$ with
    $\Sigma_{\vz\mid\vbeta}=\bigl(\begin{smallmatrix} I_d & \tilde\vbeta\\ \tilde\vbeta^\top & \tau(\vbeta)\end{smallmatrix}\bigr)$, and the query $\tilde\vx_{L+1}\sim\mathcal N(0,I_d)$ is independent of $\vz_{\mathrm c}$.
    Hence, under the pool or under the prior, the density of $\vz_{\mathrm c}$ depends on $\vz_{\mathrm c}$ only through $M$, and it is invariant under $\vz_{\mathrm c}\mapsto Q\vz_{\mathrm c}$ for every $Q\in O(L)$ acting on context positions.
    A rotation-invariant kernel is invariant under the same map applied to both arguments, since it leaves $\vz^\top\vz'=\Tr(\vz_{\mathrm c}^\top \vz_{\mathrm c}')+\vx_{L+1}^\top\vx_{L+1}'$ and $\|\vz\|$ unchanged.
    It follows that:
    \begin{enumerate}
        \item The kernel's eigenfunctions can be chosen to be either $O(L)$-invariant or orthogonal to every $O(L)$-invariant function.
        The target is invariant, so only invariant eigenfunctions carry nonzero eigencoefficients, under the pool or the prior.
        The invariant polynomials of $\vz_{\mathrm c}$ are exactly the polynomials in the entries of $M$.
        \item Within the subkernel expansion (\cref{def:see}), the degree-$p$ eigenfunctions separate by Hermite degree in the query, because the query is Gaussian and independent of the context.
        The target is linear in $\tilde\vx_{L+1}$, so it overlaps only eigenfunctions of Hermite degree one in the query.
    \end{enumerate}
    Every eigenfunction with a nonzero eigencoefficient is therefore of the form $\phi(M)^\top\tilde\vx_{L+1}$, with $\phi$ a vector of polynomials in the entries of $M$.
    At data degree $p=2k+1$, these polynomials have degree at most $k$ in $M$.
    \label{prop:M}
\end{proposition}

\begin{proof}
    For (i), the integral operator of $K$ commutes with the unitary action of $O(L)$ on $L^2$, so each of its eigenspaces is $O(L)$-invariant and splits orthogonally into its invariant functions and their complement.
    An invariant target has zero inner product with the complement.
    That the invariant polynomials are the polynomials in $M$ is the first fundamental theorem of invariant theory for $O(L)$, applied to the $d+1$ columns of $\vz_{\mathrm c}$.
    For (ii), after the projections of \cref{def:see}, a level-$p$ feature of query degree $j$ retains only the Hermite-degree-$j$ part of its query factor.
    Hermite polynomials of different degree in $\tilde\vx_{L+1}$ are orthogonal, and they are independent of the context.
\end{proof}

In particular, sums that are invariant under permutations of the context but not under $O(L)$, such as $\sum_i y_i^4$ or $\sum_i y_i^3\tilde\vx_i$, carry no coefficient of their own; their invariant parts lie in the span of polynomials in $M$.
We call a monomial type
\begin{equation}
    \psi_{(n_W,n_S,n_A)} = W^{\otimes n_W}\otimes S^{\otimes n_S}A^{n_A}\otimes\tilde\vx_{L+1},\qquad n_W+n_S+n_A=k,
\end{equation}
a \emph{family} of degree $p=2k+1$.
For example, the degree-3 families are $S\otimes\tilde\vx_{L+1}$, $A\,\tilde\vx_{L+1}$ and $W\otimes\tilde\vx_{L+1}$, and there are six families at degree 5.
Every family carries a single factor of the query.
The query is independent of the context and has $\mathbb E[\tilde\vx_{L+1}\tilde\vx_{L+1}^\top]=I_d$, so perpendicularization (as is done in \cref{app:eigensystem}) acts on the context factor alone.
We write $\tilde\Psi_p=\tilde\Psi^{\mathrm c}_p\otimes\tilde\vx_{L+1}$, where $\tilde\Psi^{\mathrm c}_p$ stacks the context factors of all degree-$p$ families, and write $\tilde\Psi^{\mathrm c\perp}_p$ for its perpendicular version.

\begin{definition}[Colored eigenfunctions, general $\Sigma_\vx$]
    Let $\Gamma_p \equiv \mathbb E\bigl[\tilde\Psi^{\mathrm c\perp}_p\,\tilde\Psi^{\mathrm c\perp\top}_p\bigr]$ be the Gram matrix of the degree-$p$ context factors.
    It is singular because the tensor families contain repeated entries, so inverses are taken on its range.
    For any orthogonal matrix $R_p$,
    \begin{equation}
        \Phi_p \equiv R_p\,\Gamma_p^{-1/2}\,\tilde\Psi^{\mathrm c\perp}_p,\qquad
        \tilde\phi_{p,m,b} = \Phi_{p,m}\,\tilde x_{L+1,b}
        \label{eq:phi-aniso}
    \end{equation}
    are orthonormal under the training measure and span the degree-$p$ eigenfunctions allowed by \cref{prop:M}.
    Unless stated otherwise we take $R_p=I$; the kernel's own eigenbasis corresponds to the $R_p$ given below.
    \label{def:orthopoly-aniso}
\end{definition}

The families of a single degree are not in general orthogonal to one another.
At degree 3 they nearly are.
The leading-order overlap between the $S$ and $A$ families is set by the pool covariance $\mathrm{Cov}_{\tilde{\mathcal B}_{\mathrm{tr}}}(\tilde\vbeta,\|\tilde\vbeta\|^2)=\Tr_1\tilde\kappa_3+2\tilde\Sigma\tilde\mu$.
For a centered prior this is $O(T^{-1/2})$, and it vanishes for a centered, symmetric pool.
The overlaps of $W$ with $S$ and $A$ fall off as $L^{-1/2}$.
From degree 5 on, each family's context factor concentrates on $L^k$ times its task object $\tilde\vbeta^{\otimes n_S}\tau^{n_A}$ (\cref{app:tasklearning}), and the pool covariances between these task objects do not vanish as $L\to\infty$.
No choice of families is then orthogonal, and the choice of $R_p$ matters.

\paragraph{The kernel's eigenbasis within a degree.}
\Cref{def:orthopoly-aniso} fixes the degree-$p$ span but not the basis within it; the kernel's eigenfunctions correspond to one particular $R_p$.
By \cref{def:see}, the subkernel $K_p=\frac{c_p}{p!}(\vz^\top\vz')^p$ acts through the degree-$p$ monomials of the raw context $\{(\vz)_p\}$, with lower degrees projected out; write $\widetilde{(\vz)}_{p,\mu}$ for these projected monomials.
By \cref{prop:M}, the span of \cref{eq:phi-aniso} is invariant under $K_p$, so step 6 of \cref{def:see} can be carried out inside it.
In the basis $\tilde\phi_{p,m,b}$ (with $R_p=I$), $K_p$ is the matrix
\begin{equation}
    [\mathcal K_p]_{(m,b),(m',b')} = \frac{c_p}{p!}\sum_\mu \mathbb E\bigl[\tilde\phi_{p,m,b}\,\widetilde{(\vz)}_{p,\mu}\bigr]\,\mathbb E\bigl[\tilde\phi_{p,m',b'}\,\widetilde{(\vz)}_{p,\mu}\bigr],
    \label{eq:kernel-rotation}
\end{equation}
which measures how strongly two orthonormal polynomials load onto the same monomials.
Its eigenvectors give the rows of $R_p$, and its eigenvalues are the kernel eigenvalues.
Because the query is independent of the context, $\mathcal K_p$ factorizes into a context part times $\Sigma_\vx$: the query index is rotated into the eigenbasis of $\Sigma_\vx$, and $R_p$ acts on the context index $m$ alone.
Since the monomials are in raw coordinates, $R_p$ depends on $\Sigma_\vx$ even though the span does not.

\subsection{The eigencoefficients}
\label{appsubsec:eigencoeffs}

We start with a lemma for evaluating conditional expectations of the context factors.

\begin{lemma}[Conditional moments of a context factor]
    For integers $a,b\ge0$, the context factor $F_{a,b}\equiv\sum_{i=1}^{L}y_i^{a}\tilde{\vx}_i^{\otimes b}$ is a sum of $L$ i.i.d.\ terms given $\vbeta$.
    Every joint cumulant of such factors given $\vbeta$ is therefore $L$ times the corresponding joint cumulant at a single position; in particular
    \begin{equation}
        \mathbb{E}\big[F_{a,b} \,\big|\, \vbeta\big] = L\, \mathbb{E}\big[y^{a}\tilde{\vx}^{\otimes b} \,\big|\, \vbeta\big],
        \qquad
        \mathrm{Cov}\big(F_{a,b} \,\big|\, \vbeta\big) = L\, \mathrm{Cov}\big(y^{a}\tilde{\vx}^{\otimes b} \,\big|\, \vbeta\big).
        \label{eq:factor-moments}
    \end{equation}
    The single-position cases used below are
    \begin{align}
        \mathbb{E}\big[y \tilde{\vx} \,\big|\, \vbeta\big] &= \tilde{\vbeta},
        &\mathrm{Cov}\big(y \tilde{\vx} \,\big|\, \vbeta\big) &= \tilde{\vbeta}\tilde{\vbeta}^\top + \tau I_d,
        \label{eq:mom-yx}\\
        \mathbb{E}\big[y^2 \,\big|\, \vbeta\big] &= \tau,
        &\mathrm{Var}\big(y^2 \,\big|\, \vbeta\big) &= 2\tau^2,
        \label{eq:mom-y2}\\
        \mathbb{E}\big[\tilde x_a\tilde x_c \,\big|\, \vbeta\big] &= \delta_{ac},
        &\mathrm{Cov}\big(\tilde x_a\tilde x_c,\, \tilde x_{a'}\tilde x_{c'} \,\big|\, \vbeta\big) &= \delta_{aa'}\delta_{cc'}+\delta_{ac'}\delta_{ca'},
        \label{eq:mom-xx}\\
        \mathrm{Cov}\big(y\tilde\vx,\,y^2\,\big|\, \vbeta\big) &= 2\tau\tilde\vbeta,
        &\mathrm{Cov}\big(y\tilde x_b,\, \tilde x_a\tilde x_c \,\big|\, \vbeta\big) &= \delta_{ab}\tilde\beta_c+\delta_{bc}\tilde\beta_a,
        \label{eq:mom-cross}\\
        \mathrm{Cov}\big(y^2,\, \tilde x_a\tilde x_c \,\big|\, \vbeta\big) &= 2\tilde\beta_a\tilde\beta_c ,
        \nonumber
    \end{align}
    with $\tau=\tau(\vbeta)$.
    \label{lem:context-moments}
\end{lemma}

\begin{proof}
    The positions are independent given $\vbeta$, and cumulants add over independent summands.
    For \eqref{eq:mom-yx}, $y\tilde{\vx} = (\tilde{\vbeta}^\top\tilde{\vx})\tilde{\vx} + \eta\tilde{\vx}$ has mean $\tilde{\vbeta}$, and by Isserlis' theorem $\mathbb{E}[(\tilde{\vbeta}^\top\tilde{\vx})^2 \tilde{\vx}\tilde{\vx}^\top] = 2\tilde{\vbeta}\tilde{\vbeta}^\top + \|\tilde{\vbeta}\|^2 I_d$, so the second moment is $2\tilde{\vbeta}\tilde{\vbeta}^\top + \tau I_d$ and the covariance follows.
    For \eqref{eq:mom-y2}, $y\,|\,\vbeta \sim\mathcal{N}(0,\tau)$ gives $\mathrm{Var}(y^2\,|\,\vbeta)=2\tau^2$.
    The remaining entries follow in the same way, since $(\tilde\vx,y)\,|\,\vbeta$ is jointly Gaussian with covariance $\Sigma_{z\mid\vbeta}$ (\cref{prop:M}).
\end{proof}

\paragraph{Task objects.}
By \cref{lem:context-moments}, each context sum concentrates on a function of the task: $W\approx L I_d$, $S\approx L\tilde\vbeta$ and $A\approx L\tau$.
The context factor of a degree-$p=2k+1$ family therefore has conditional mean
\begin{equation}
    \mathbb E\bigl[W^{\otimes n_W}\otimes S^{\otimes n_S}A^{n_A}\,\big|\,\vbeta\bigr]
    = L^{k}\, I_d^{\otimes n_W}\otimes\tilde\vbeta^{\otimes n_S}\tau^{n_A} + O(L^{k-1}),
\end{equation}
where the $O(L^{k-1})$ corrections come from coincident positions.
We call $f(\vbeta)=I_d^{\otimes n_W}\otimes\tilde\vbeta^{\otimes n_S}\tau^{n_A}$ the family's \emph{task object}.
We write $\vf_p$ for the stacked task objects of all degree-$p$ families, $\vf_p^\perp$ for their residual after regression on all lower-degree task objects under the pool, and
\begin{equation}
    G^\perp_p \equiv \mathrm{Cov}_{\tilde{\mathcal B}_{\mathrm{tr}}}\bigl(\vf_p^\perp\bigr).
\end{equation}
Directions with $G_p^\perp\neq0$ are \emph{task modes}.
Directions with $G_p^\perp=0$ are \emph{noise modes}: the $W$ families, and trace components such as $\Tr(S\otimes S)=\|S\|^2$, whose task objects repeat lower-degree ones.

\begin{theorem}[The eigencoefficient-cumulant correspondence]
    Let $\tilde\phi_{p,m,b}=\Phi_{p,m}\tilde x_{L+1,b}$ be the orthonormal polynomials of \cref{def:orthopoly-aniso}, and let $\vq_p(\vbeta)\equiv\mathbb E\bigl[\tilde\Psi^{\mathrm c\perp}_p\,\big|\,\vbeta\bigr]$.
    \begin{enumerate}
        \item \emph{Exact form.} The eigencoefficients are $v_{1,b}=\tilde\mu_b$ and, for $p\ge3$,
        \begin{equation}
            v_{p,\cdot,b}=R_p\,\Gamma_p^{-1/2}\,V_{p,b},
            \qquad
            V_{p,b}=\mathbb E\bigl[\tilde\beta_b\,\tilde\Psi^{\mathrm c\perp}_p\bigr]
            =\mathrm{Cov}_{\tilde{\mathcal B}_{\mathrm{tr}}}\bigl(\tilde\beta_b,\ \boldsymbol q_p(\vbeta)\bigr).
            \label{eq:v-aniso}
        \end{equation}
        \item \emph{Scaling with $L$.} Degree-3 task-mode coefficients are $O(1)$. For $p=2k+1\ge5$, $V_{p,b}=O(L^{k-1})$ while $\Gamma_p=L^{2k}G_p^\perp+O(L^{2k-1})$, so task-mode coefficients are $O(L^{-1})$ and noise-mode coefficients are $O(L^{-1/2})$.
        \item \emph{Cumulant content.} At degree $2k+1$, the numerator of a family has at most $k+1$ free task indices, and only $S^{\otimes k}$ attains this bound. The cumulant $\tilde\kappa_\omega$ therefore first appears with all indices free at degree $2\omega-1$, in the $S^{\otimes(\omega-1)}$ block; every lower degree contains it only through partial traces. The diagonal block of $\Gamma_p$ for $S^{\otimes(\omega-1)}$ involves pool moments up to order $2\omega-2$, and the full $\Gamma_p$, through the $A$ families, involves moments up to order $4\omega-4$.
    \end{enumerate}
    \label{thm:cum-aniso}
\end{theorem}

\begin{proof}
    \emph{(i).} By \cref{lem:kernel-memorize}, $v_{p,m,b}=\mathbb E[y^\star\tilde\phi_{p,m,b}]$.
    The query is independent of the context and $\mathbb E[y_{L+1}\tilde\vx_{L+1}\,|\,\vbeta]=\tilde\vbeta$, so $v_{p,m,b}=\mathbb E[\tilde\beta_b\,\Phi_{p,m}]$, and \cref{eq:phi-aniso} gives the first equality.
    Conditioning on $\vbeta$ gives $V_{p,b}=\mathbb E_{\tilde{\mathcal B}_{\mathrm{tr}}}[\tilde\beta_b\vq_p]$.
    For $p\ge3$, $\tilde\Psi^{\mathrm c\perp}_p$ is orthogonal to the constant (the degree-1 context factor), so $\mathbb E[\vq_p]=0$ and this expectation is a covariance.
    For $p=1$ the context factor is $1$, and $v_{1,b}=\tilde\mu_b$.

    \emph{(ii).} By \cref{lem:context-moments}, the covariance between context factors of degrees $2k+1$ and $2k'+1$ is $L^{k+k'}$ times the pool covariance of their task objects, up to $O(L^{k+k'-1})$.
    At leading order, the perpendicularization of \cref{def:perppoly} is therefore the regression of $\vf_p$ on the lower-degree task objects under the pool, so $\vq_p=L^k\vf_p^\perp+O(L^{k-1})$ and $\Gamma_p=L^{2k}G_p^\perp+O(L^{2k-1})$.
    At $p=3$, the only lower-degree task object is the constant, so $V_{3,b}=L\,\mathrm{Cov}(\tilde\beta_b,\boldsymbol f_3)+O(1)$ and the task-mode coefficients are $O(1)$.
    For $p\ge5$, the lower degrees include the $S$ family, whose task object is $\tilde\vbeta$ itself.
    The residual $\boldsymbol f_p^\perp$ is then uncorrelated under the pool with every affine function of $\tilde\vbeta$, so $\mathrm{Cov}(\tilde\beta_b,\boldsymbol f_p^\perp)=0$ and the $O(L^k)$ part of $V_{p,b}$ vanishes.
    What survives is $O(L^{k-1})$.
    It arises because the data-space regression is onto the context factors rather than onto their task objects.
    For instance, $S$ has conditional covariance $L(\tilde\vbeta\tilde\vbeta^\top+\tau I_d)$ by \eqref{eq:mom-yx}, which attenuates its regression coefficient by a relative $O(L^{-1})$ and leaves part of the linear component of $\boldsymbol f_p$ in place.
    Dividing by $\Gamma_p^{1/2}$, which is $O(L^k)$ on task modes and $O(L^{k-1/2})$ on noise modes, gives the stated orders.

    \emph{(iii).} A factor $W$ contributes $I_d$ and a factor $A$ contributes the scalar $\tau$, so the task object of $(n_W,n_S,n_A)$ has $n_S$ free task indices, and the target adds the index $b$.
    Contracting against $I_d$, $\tau$ or the regression coefficients adds no free indices, so $V_{p,b}$ has at most $n_S+1\le k+1$ of them, with equality only for $S^{\otimes k}$.
    A pool moment of order $\omega$ with all of its indices free, and hence $\tilde\kappa_\omega$, requires $k+1\ge\omega$.
    It can therefore first appear at degree $2\omega-1$, and only in the $S^{\otimes(\omega-1)}$ block.
    It does appear there: the attenuation term of (ii) contains a term proportional to $\mathrm{Cov}_{\tilde{\mathcal B}_{\mathrm{tr}}}(\tilde\vbeta^{\otimes(\omega-1)},\tilde\vbeta)$, as in the example below.
    The diagonal block of $\Gamma_p$ for $S^{\otimes(\omega-1)}$ pairs two task objects of degree $\omega-1$, giving moments up to order $2\omega-2$.
    The $A$ families have task objects of degree up to $2\omega-2$ in $\tilde\vbeta$, so the full $\Gamma_p$ involves moments up to order $4\omega-4$.
\end{proof}

The above eigencoefficient equation is highly opaque, and will remain so for full completeness.
Nonetheless, we detail the first few eigencoefficients for lower-degree polynomials.

\paragraph{Example: the $SS$ block at degree 5.}
Take a centered pool with $\Tr_1\tilde\kappa_3=0$.
Then $S$ is uncorrelated with $A$, $W$ and the constant, and only the regression of $S\otimes S$ on $S$ contributes to $V_{SS,b}$.
By \cref{lem:context-moments},
\begin{align*}
    \mathbb E[S\otimes S\,|\,\vbeta]&=L(L+1)\,\tilde\vbeta\tilde\vbeta^\top+L\tau I_d,\qquad
    \mathrm{Cov}(S)=L(L+1)\tilde\Sigma+L\bar\tau I_d\\
    \mathbb E[S^{\otimes3}]&=(L^3+3L^2)\,\tilde\kappa_3+O(L).
\end{align*}
The regression coefficient of $S\otimes S$ on $S$ is $\mathbb E[S^{\otimes3}]\,\mathrm{Cov}(S)^{-1}$, and $\mathrm{Cov}(\tilde\beta_b,\mathbb E[S\,|\,\vbeta])=L\tilde\Sigma_{b\cdot}$, so
\begin{equation*}
    V_{SS,b}=L(L+1)\,\tilde\kappa_{3,b\cdot\cdot}
    -(L^3+3L^2)\bigl[\bigl((L+1)\tilde\Sigma+\bar\tau I_d\bigr)^{-1}\tilde\Sigma\bigr]_{bb'}\tilde\kappa_{3,b'\cdot\cdot}
    =L\bigl[(\bar\tau\tilde\Sigma^{-1}-I_d)\,\tilde\kappa_3\bigr]_{b\cdot\cdot}+O(1).
\end{equation*}
The $O(L^2)$ terms cancel as in (ii).
The surviving $\bar\tau\tilde\Sigma^{-1}$ is the per-position noise-to-signal ratio of $S$, which sets the attenuation.
With $\Gamma_{SS}=L^4G_5^\perp+O(L^3)$, this gives the $SS$ entry of \cref{apptable:eigcoeffs}.

\Cref{eq:v-aniso} is exact, but its content is easiest to read family by family.
\Cref{apptable:eigcoeffs} lists the leading-order coefficients through degree seven.

\begin{table}[h]
\centering
\small
\begin{tabular}{clll}
\toprule
Degree $p$ & family & task object & eigencoefficient (leading order) \\
\midrule
$1$ & $1$ & $1$ & $\tilde\mu_{b}$ \\
\midrule
$3$ & $S$ & $\tilde\vbeta$ & $\tilde s_a/\sqrt{(1+L^{-1})\tilde s_a+\bar\tau/L}$ \\
    & $A$ & $\tau$ & $c_b/\sqrt{\mathrm{Var}\,\tau+2\,\mathbb E\tau^2/L}$,\quad $c=\Tr_1\tilde\kappa_3+2\tilde\Sigma\tilde\mu$ \\
    & $W$ & $I_d$ & $0$ (noise mode) \\
\midrule
$5$ & $SS$ & $\tilde\vbeta^{\otimes2}$ & $L^{-1}(G_5^\perp)^{-1/2}(\bar\tau\tilde\Sigma^{-1}-I_d)\tilde\kappa_3$ \\
    & $SA$ & $\tau\tilde\vbeta$ & $L^{-1}(G_5^\perp)^{-1/2}\left(-(2\Tr\tilde\Sigma^2\,I_d+4\tilde\Sigma^2)+\dots\right)$ \\
    & $AA$ & $\tau^{2}$ & $O(L^{-1})$; odd moments \\
    & $W$ families & $\dots$ & $O(L^{-1/2})$ (noise modes) \\
\midrule
$7$ & $SSS$ & $\tilde\vbeta^{\otimes3}$ & $L^{-1}(G_7^\perp)^{-1/2}\tfrac{d}{d+2}\tfrac{\epsilon^2}{\tilde s}\tilde\kappa_4$ \\
    & $SAA$ & $\tau^2\tilde\vbeta$ & $\propto L^{-1}\,\Tr_2\tilde\kappa_6$ \\
    & $SSA$, $AAA$ & $\tau\tilde\vbeta^{\otimes2}$, $\tau^3$ & odd moments; zero for symmetric pools \\
    & $W$ families & lower-degree objects & $O(L^{-1/2})$ (noise modes) \\
\bottomrule
\end{tabular}
    \caption{Leading-order eigencoefficients through degree seven, in the symmetric basis $R_p=I$ (\cref{def:orthopoly-aniso}).
    $G_p^\perp$ is the pool covariance of the degree-$p$ task objects after regression on lower-degree ones, joint across families.
    Degree-3 rows treat the $S$, $A$ and $W$ blocks as uncoupled and are written in the eigenbasis of $\tilde\Sigma$.
    The $S$ row is exact at every $L$ when $\tilde\mu=0$ and $\Tr_1\tilde\kappa_3=0$, and the $A$ row is first order in $c$.
    Degree-5 rows are leading order in $1/L$. The $SS$ row assumes $\tilde\mu=0$ and $\Tr_1\tilde\kappa_3=0$; otherwise terms in $\tilde\mu$ and $\Tr_1\tilde\kappa_3$ are added.
    The $SA$ row shows the Gaussian part for a centered pool.
    Degree-7 rows are leading order in $1/L$ for an isotropic centered pool ($\tilde\Sigma=\tilde s\,I_d$), at first order in the deviation from a Gaussian.
    Noise modes are the $W$ families and trace components such as $\Tr(S\otimes S)$ (part of the $SS$ family), whose task objects repeat lower-degree ones.}
\label{apptable:eigcoeffs}
\end{table}

\subsection{Validation of the eigencoefficients}

\begin{figure}[t]
    \includegraphics[width=\textwidth]{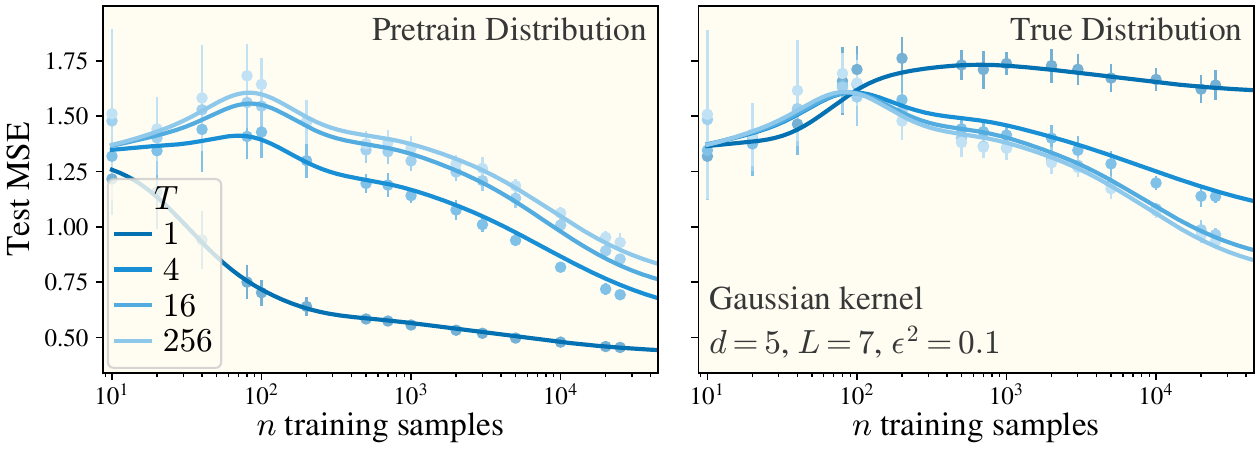}
    \caption{
    \textbf{Gaussian kernel learning curves on ICL Regression data.}
    For four different settings (colors) of the task diversity $T$, we compute the expected test MSE as a function of the number of training samples $n$, and compare it to empirical errors obtained across $20$ trials.
    % \remind \joey{True to prior distribution}
    Theoretical learning curve equations are provided in \cref{app:krr_eigenframework}, with validation of the eigensystem the equations are built off of provided in \cref{app:eigensystem,app:correspondence}.
    Our data uses a Gaussian kernel with kernel width $\sigma_\text{ker}=\sqrt3$, $d=5, L=7, \epsilon^2=0.1$ where our data is drawn $\vx \sim \mathcal{N}(0, I_d/(dL))$.
    % Between the test MSE on tasks drawn from the pretraining distribution (left) and novel tasks specifying the true distribution (right), we highlight that for $T<T'$, the $\text{test MSE}_{\text{pretrain}}(T) > \text{test MSE}_{\text{true}}(T)$, which is shown by the positions of the curves being inverted between the pretrain and true distributions.
    Additionally, as $T\to \infty$, we find that the errors on the pretraining distribution match those of the true distribution, as expected.
    Further experimental details are in \cref{app:experiment_details}.
    }
    \label{fig:learning_curves}
\end{figure}

To validate our theoretically derived eigencoefficients as the correct set used by the kernel learner, we use the theoretically-derived learning curve (test error vs number of samples) plotted against empirically found errors.
We do not directly compare the eigencoefficients that are found empirically to those theoretically derived, as we note that the ``interesting region'' of the eigencoefficients (the portion of the spectrum beyond the degree 1 terms) is partially affected by noise, with us not having enough memory to obtain clean empirical estimates.
Therefore, we test our eigencoefficients in a different manner.
Should our eigensystem, including the eigencoefficients, be correct, the learning curve (test error given number of samples) can be theoretically predicted, and will match well with empirical findings.
In \cref{fig:learning_curves}, we do said experiment and find excellent agreement between our pure-theory curves and the empirically found kernel regression errors.

\section{Task space learning}
\label{app:tasklearning}

In \cref{app:eigensystem,app:correspondence}, we described the kernel's eigensystem in the raw data space of $Z$.
The same results can be read in \emph{task space}: averaging the predictor over the in-context inputs and noise, $\vx$ and $\eta$, gives the task the kernel predicts on average for a given latent $\vbeta$.
Within the eigenframework this average is exact; the approximations below enter only through the large-$L$ forms of the task images.

\subsection{Eigenfunctions in task space}
\label{app:task_eigenfunctions}

Write each orthonormal polynomial of \cref{def:orthopoly-aniso} as $\tilde\phi_{p,m,b}(Z)=\Phi_{p,m}\,\tilde x_{L+1,b}$, where $\Phi_{p,m}$ depends only on the context, and define its \emph{task image}
\begin{equation}
    \chi_{p,m}(\vbeta)\equiv\mathbb{E}_{\tilde{\vx},\eta}\bigl[\Phi_{p,m}\,\big|\,\vbeta\bigr],
    \qquad\text{so that}\qquad
    \boldsymbol\chi_p(\vbeta)=R_p\,\Gamma_p^{-1/2}\,\boldsymbol q_p(\vbeta),
    \label{eq:chi-def}
\end{equation}
with $\boldsymbol q_p$ as in \cref{thm:cum-aniso}.
By \cref{thm:cum-aniso}(i), the eigencoefficients are task-space overlaps.
By the law of total covariance, the task images are orthonormal up to the within-context noise:
\begin{equation}
    v_{p,m,b}=\mathbb E_{\tilde{\mathcal B}_{\mathrm{tr}}}\big[\tilde\beta_b\,\chi_{p,m}\big],\qquad
    \mathbb E_{\tilde{\mathcal B}_{\mathrm{tr}}}\big[\chi_{p,m}\chi_{p',m'}\big]=\delta_{pp'}\delta_{mm'}-\mathbb E_{\tilde{\mathcal B}_{\mathrm{tr}}}\big[\mathrm{Cov}(\Phi_{p,m},\Phi_{p',m'}\,|\,\vbeta)\big].
    \label{eq:v-chi}
\end{equation}

At large $L$, the task images simplify.
By \cref{thm:cum-aniso}(ii), $\boldsymbol q_p=L^k\boldsymbol f_p^\perp+O(L^{k-1})$ and $\Gamma_p=L^{2k}G_p^\perp+O(L^{2k-1})$, with the task objects $\boldsymbol f_p$ and their pool covariance $G_p^\perp$ as defined in \cref{appsubsec:eigencoeffs}.
On task modes, therefore,
\begin{equation}
    \boldsymbol\chi_p(\vbeta)\approx R_p\,(G_p^\perp)^{-1/2}\,\boldsymbol f_p^\perp(\vbeta),
    \label{eq:chi-largeL}
\end{equation}
while on noise modes $\boldsymbol f_p^\perp=0$ and the task images vanish as $O(L^{-1/2})$.
At large $L$, the task images are thus orthonormal polynomials of $\tilde\vbeta$ under the pool measure.
At degree $p=2k+1$, the new objects are $\tilde\vbeta^{\otimes n_S}\tau^{n_A}$ with $n_S+n_A=k$, of degree $k$ to $2k$ in $\tilde\vbeta$.
As in data space, each degree is regressed on all lower degrees and then orthonormalized as a block through $G_p^\perp$.

\subsection{Eigencoefficients in task space}

Write the predictor as $f(Z)=\hat{\tilde\vbeta}{}^\top\tilde\vx_{L+1}$, where $\hat{\tilde\vbeta}$ is the predicted colored task.
Take $\Sigma_\vx\propto I_d$, as in all our experiments, so the learnability $\mathcal L_{p,m}$ of a mode does not depend on the query index.
Averaging over the context given $\vbeta$ gives
\begin{equation}
    \mathbb{E}_{\tilde{\vx},\eta}\bigl[\hat{\tilde\beta}_b\,\big|\,\vbeta\bigr] = \sum_{p,m} \mathcal{L}_{p,m}\,v_{p,m,b}\,\chi_{p,m}(\vbeta).
    \label{eq:betahat-task}
\end{equation}
Here $v$ and $\chi$ must be taken in the kernel's own basis (\cref{eq:kernel-rotation}), since that is the basis in which learnabilities are assigned.
For general $\Sigma_\vx$, $\mathcal L$ also depends on the query index through the eigenvalues of $\Sigma_\vx$, with $b$ taken in that eigenbasis.
Once a degree is fully learned ($\mathcal L_{p,m}=1$ for all $m$), its contribution is basis-independent:
\begin{equation}
    \sum_m v_{p,m,b}\,\chi_{p,m}(\vbeta)=V_{p,b}^\top\,\Gamma_p^{-1}\,\boldsymbol q_p(\vbeta)
    \;\approx\; L^{-k}\,V_{p,b}^\top\,(G_p^\perp)^{-1}\boldsymbol f_p^\perp(\vbeta).
    \label{eq:degree-contrib}
\end{equation}

The terms arrive in order of degree.
At degree 1, the kernel learns the pool mean $\tilde\mu$.
At degree 3, for $\tilde\mu=0$ and $c=\Tr_1\tilde\kappa_3+2\tilde\Sigma\tilde\mu=0$, the $S$ modes contribute
\begin{equation*}
    \sum_a\frac{L\tilde s_a}{(L+1)\tilde s_a+\bar\tau}\,\vu_a\vu_a^\top\tilde\vbeta ,
\end{equation*}
a shrinkage estimate of the true task.
It recovers $\tilde\vbeta$ in directions where $\tilde s_a\gg\bar\tau/L$, and not at all in directions the pool does not span ($\tilde s_a=0$, which is unavoidable when $T\le d$).
A nonzero $c$ adds the $A$ mode, a scalar function of $\tau(\vbeta)$ along the fixed direction $c$.
From degree 5 on, the learner adds nonlinear functions of $\tilde\vbeta$ of degree $k$ to $2k$, each with coefficient $O(L^{-1})$.
By \cref{thm:cum-aniso}(iii), these bring in the pool's skewness at degree 5, its kurtosis at degree 7, and so on.
\Cref{apptable:task-channels} lists the task images and coefficients through degree seven.

\begin{table}[ht]
\centering
\small
\begin{tabular}{ccll}
\toprule
Degree $p$ & family & $\chi_{p,m}(\vbeta)$ & $v_{p,m,b}$ \\
\midrule
$1$ & $1$ & $1$ & $\tilde\mu_{b}$\\[4pt]
\midrule
$3$ & $S$ & $\dfrac{\vu_a^\top\tilde{\vbeta}}{\sqrt{(1+L^{-1})\tilde s_a+\bar\tau/L}}$
    & $\dfrac{\tilde s_a}{\sqrt{(1+L^{-1})\tilde s_a+\bar\tau/L}}$\\[10pt]
$3$ & $A$ & $\dfrac{\tau-\bar\tau}{\sqrt{\mathrm{Var}\,\tau+2\,\mathbb E\tau^2/L}}$
    & $\dfrac{c_b}{\sqrt{\mathrm{Var}\,\tau+2\,\mathbb E\tau^2/L}}$\\[10pt]
$3$ & $W$ & $0$ & $0$ (noise mode)\\[4pt]
\midrule
$5$ & $SS$ & $(G_5^\perp)^{-1/2}\big(\tilde{\vbeta}^{\otimes2}\big)^{\perp}$
    & $L^{-1}(G_5^\perp)^{-1/2}(\bar\tau\tilde\Sigma^{-1}-I_d)\tilde\kappa_3$\\[4pt]
$5$ & $SA$ & $(G_5^\perp)^{-1/2}\big(\tau\tilde{\vbeta}\big)^{\perp}$
    & $L^{-1}(G_5^\perp)^{-1/2}\left(-(2\Tr\tilde\Sigma^2\,I_d+4\tilde\Sigma^2)+\dots\right)$\\[4pt]
$5$ & $AA$ & $(G_5^\perp)^{-1/2}\big(\tau^{2}\big)^{\perp}$
    & $O(L^{-1})$; odd moments\\[4pt]
$5$ & $W$ families & $O(L^{-1/2})$ & $O(L^{-1/2})$ (noise modes)\\[4pt]
\midrule
$7$ & $SSS$ & $(G_7^\perp)^{-1/2}\big(\tilde{\vbeta}^{\otimes3}\big)^{\perp}$
    & $L^{-1}(G_7^\perp)^{-1/2}\tfrac{d}{d+2}\tfrac{\epsilon^2}{\tilde s}\tilde\kappa_4$\\[4pt]
$7$ & $SSA$ & $(G_7^\perp)^{-1/2}\big(\tau\tilde{\vbeta}^{\otimes2}\big)^{\perp}$
    & odd moments\\[4pt]
$7$ & $SAA$ & $(G_7^\perp)^{-1/2}\big(\tau^{2}\tilde{\vbeta}\big)^{\perp}$
    & $\propto L^{-1}\,\Tr_2\tilde\kappa_6$\\[4pt]
$7$ & $AAA$ & $(G_7^\perp)^{-1/2}\big(\tau^{3}\big)^{\perp}$
    & odd moments\\[4pt]
$7$ & $W$ families & $O(L^{-1/2})$ & $O(L^{-1/2})$ (noise modes)\\[4pt]
\bottomrule
\end{tabular}
    \caption{Task images and eigencoefficients through degree seven, in the symmetric basis $R_p=I$.
    $(\cdot)^{\perp}$ is the regression on all lower-degree task objects under the pool measure, and $(G_p^\perp)^{-1/2}$ is the symmetric inverse square root of the joint degree-$p$ pool covariance.
    The family column therefore labels blocks of $\boldsymbol f_p^\perp$, which this rotation mixes unless $G_p^\perp$ is block diagonal.
    A family with $n_S$ factors of $S$ has a task image of tensor rank $n_S$; its coefficient carries $n_S+1$ indices, of which $n_S$ contract against the task image and one is the output index $b$.
    Degree-3 rows are exact at every $L$ and are written in the eigenbasis of $\tilde\Sigma$: the $S$ row when $\tilde\mu=0$ and $\Tr_1\tilde\kappa_3=0$, and the $A$ row at first order in $c=\Tr_1\tilde\kappa_3+2\tilde\Sigma\tilde\mu$.
    Degree-5 rows are leading order in $1/L$; the $SS$ row assumes $\tilde\mu=0$ and $\Tr_1\tilde\kappa_3=0$, and the $SA$ row shows the Gaussian part for a centered pool.
    Degree-7 rows are leading order in $1/L$ for an isotropic centered pool ($\tilde\Sigma=\tilde s\,I_d$), at first order in the deviation from a Gaussian. $\chi$ is quoted at $L\to\infty$ except at degree 3.}
\label{apptable:task-channels}
\end{table}

\section{Oracle expansions}
\label{app:oracles}

Throughout this appendix we work in raw coordinates, with $W \equiv \sum_{i=1}^L \vx_i\vx_i^\top$, $S \equiv \sum_{i=1}^L y_i\vx_i$, $A\equiv\sum_{i=1}^L y_i^2$, and $\Delta W \equiv W - L\Sigma_\vx$.
These are the unwhitened versions of the sums in \cref{appsubsec:whitened}.
We write $\Sigma_\vbeta$ and $\vmu_\vbeta$ for the prior's covariance and mean, and $\hat\mu_\vbeta,\hat\Sigma_\vbeta,\hat\kappa_\ell$ for the cumulants of the pool $\mathcal{B}_{\mathrm{tr}}$.

Both oracles are Bayes-optimal under the \emph{same} likelihood, $y_i \mid \vbeta \sim \mathcal{N}(\vbeta^\top\vx_i,\epsilon^2)$; they differ only in the prior they place on $\vbeta$.
The ridge takes the population Gaussian $\mathcal{N}(\vmu_\vbeta,\Sigma_\vbeta)$, and the dMMSE takes the uniform measure on the finite pool $\mathcal{B}_{\mathrm{tr}}$.
Everything in this appendix follows from that single difference.

\subsection{The generalizing solution}
\label{app:ridge}

For $\vbeta\sim\mathcal{N}(\vmu_\vbeta,\Sigma_\vbeta)$, the Bayes-optimal predictor is $\hat f_{\mathrm{ridge}}=\vbeta_{\mathrm{ridge}}^\top\vx_{L+1}$, where $\vbeta_{\mathrm{ridge}}$ is ridge regression with matrix parameter $\epsilon^2\Sigma_\vbeta^{-1}$:
\begin{equation}
    \vbeta_{\mathrm{ridge}} = \left(W + \epsilon^2\Sigma_\vbeta^{-1}\right)^{-1}\left(S+\epsilon^2\Sigma_\vbeta^{-1}\vmu_\vbeta\right).
    \label{eq:ridge}
\end{equation}
The main text takes $\vmu_\vbeta=0$; we keep the general form for \cref{appsubsec:dmmsevsridge}.

\subsection{The memorizing solution}
\label{app:dmmse}

For $\vbeta$ uniform on $\mathcal B_{\mathrm{tr}}$, the Bayes-optimal predictor is $\hat f_{\mathrm{mem}}=\hat\vbeta_{\mathrm{mem}}^\top\vx_{L+1}$, with the posterior mean over the pool
\begin{equation}
    \hat\vbeta_{\mathrm{mem}}=\sum_{t=1}^T w_t\,\vbeta^{(t)},\qquad
    w_t=\frac{e^{\ell_t}}{\sum_{s}e^{\ell_s}},\qquad
    \ell_t=\frac{1}{\epsilon^2}\Bigl(\vbeta^{(t)\top}S-\tfrac12\,\vbeta^{(t)\top}W\vbeta^{(t)}\Bigr).
    \label{eq:dmmse-weights}
\end{equation}
This is \cref{eq:memdef}, after expanding $\sum_i(y_i-\vbeta^\top\vx_i)^2=A-2\vbeta^\top S+\vbeta^\top W\vbeta$ and cancelling $A$, which is common to every task.
Two consequences are used below.
First, both oracles depend on the context only through $W$ and $S$; the sum $A$ enters the eigencoefficients only through its correlation with functions of $W$ and $S$.
Second, $\hat\vbeta_{\mathrm{mem}}=\epsilon^2\nabla_S\log\sum_te^{\ell_t}$.
A Taylor series of the dMMSE in $S$ therefore generates cumulants of the pool reweighted by $e^{-\vbeta^\top W\vbeta/2\epsilon^2}$, which at $W\approx L\Sigma_\vx$ is strongly tilted, rather than cumulants of the pool itself.
The pool's own cumulants enter only after projection onto the orthonormal basis (\cref{thm:cum-aniso}).

\subsection{Why no finite truncation of the oracle expansions yields the eigencoefficients}
\label{appsubsec:expansion}

It is tempting to skip choosing the eigenfunction basis and instead Taylor-expand the oracles directly.
This does not work: reasonable expansions identify the modes an oracle occupies, but truncating at any finite order misstates their coefficients, by arbitrary factors and in some cases all the way to zero.
We record two examples as a warning.

\paragraph{Ridge.}
With $r_j\equiv y_j-\vmu_\vbeta^\top\vx_j$ and $\mP \equiv \left(L\Sigma_\vx + \epsilon^2\Sigma_{\vbeta}^{-1}\right)^{-1}$, so that $W+\epsilon^2\Sigma_\vbeta^{-1}=\mP^{-1}+\Delta W$, the Neumann series gives
\begin{align}
    \hat f_{\mathrm{ridge}}
    &= \vmu_\vbeta^\top\vx_{L+1} + \vx_{L+1}^\top\left(W+\epsilon^2\Sigma_\vbeta^{-1}\right)^{-1}\sum_{j=1}^L r_j\vx_j
    \nonumber\\
    &= \vmu_\vbeta^\top\vx_{L+1} + \sum_{m\ge0}(-1)^m\,\vx_{L+1}^\top\mP(\Delta W\mP)^m\sum_{j=1}^L r_j\vx_j
    \nonumber\\
    &= \vmu_\vbeta^\top\vx_{L+1} + \sum_{j} r_j\,\vx_j^\top\mP\,\vx_{L+1}\nonumber\\
    &\qquad- \Bigl(\sum_{i,j} r_j\,(\vx_j^\top\mP\vx_i)(\vx_i^\top\mP\vx_{L+1})
    - L\sum_{j} r_j\,\vx_j^\top\mP\Sigma_\vx\mP\,\vx_{L+1}\Bigr) + \dots,
    \label{eq:ridge-expansion}
\end{align}
which converges since $\|\mP\,\Delta W\|=O(L^{-1/2})$.
The first line uses $S-W\vmu_\vbeta=\sum_jr_j\vx_j$.

Take $\Sigma_\vx\propto I_d$ and $\vmu_\vbeta=0$, and evaluate under the prior, so that $(\tilde s_a,\vu_a)$ are the eigenpairs of $\Sigma_\vx^{1/2}\Sigma_\vbeta\Sigma_\vx^{1/2}$ and $\bar\tau=\Tr(\Sigma_\vx\Sigma_\vbeta)+\epsilon^2$.
Truncating \eqref{eq:ridge-expansion} at data degree three keeps $\sum_j r_j\,\vx_j^\top\mP\,\vx_{L+1}$.
The coefficient it assigns to the normalized degree-3 mode is the raw coefficient \emph{times} the mode's norm,
\begin{equation}
    v^{\mathrm{trunc}}_a
    = \frac{\tilde s_a}{L\tilde s_a+\epsilon^2}\,
      \sqrt{L(L+1)\tilde s_a+L\bar\tau}\,,
    \label{eq:v-trunc}
\end{equation}
whereas the projection gives the overlap \emph{divided} by the norm (\cref{apptable:eigcoeffs}),
\begin{equation}
    v^{\mathrm{proj}}_a
    = \frac{L\,\tilde s_a}{\sqrt{L(L+1)\tilde s_a+L\bar\tau}}
    = \frac{\tilde s_a}{\sqrt{(1+L^{-1})\tilde s_a+\bar\tau/L}}\,.
    \label{eq:v-proj}
\end{equation}
Their ratio is
\begin{equation}
    \frac{v^{\mathrm{trunc}}_a}{v^{\mathrm{proj}}_a}
    = \frac{(L+1)\tilde s_a+\bar\tau}{L\tilde s_a+\epsilon^2}
    = 1 + \frac{\tilde s_a+\bar\tau-\epsilon^2}{L\tilde s_a+\epsilon^2}
    \;>\;1 ,
    \label{eq:trunc-ratio}
\end{equation}
since $\bar\tau\ge\epsilon^2$.
The truncation therefore always overestimates, for every mode and every parameter setting, and approaches the projection only as $O(L^{-1})$.
For example, at $d=2$, $L=6$, $\epsilon^2=0.5$, $\Sigma_\vx=I_d/2$ and $\Sigma_{\vbeta}=I_d$, one has $\tilde s_a=0.5$ and $\bar\tau=1.5$, giving a ratio of $5/3.5\approx1.43$.

The two operations differ because the eigencoefficients are \emph{projections} of the oracle onto an orthonormal basis, $v_i=\mathbb E_Z[y^\star\phi_i]$, while an expansion \emph{truncates} the oracle itself.
They agree only when the discarded terms are orthogonal to the retained modes, and regular polynomials of different degree are not.
Here, every higher-order term of \eqref{eq:ridge-expansion} generally has a nonzero projection onto the degree-3 mode, so the correct coefficient requires resumming the whole series.

\paragraph{dMMSE.}
The $A := \sum_i y_i^2$ modes give a more extreme case.
By \cref{app:dmmse}, the dMMSE contains no quadratic functions of $y_i^2$ at all (equivalent factors appear in both the numerator and denominator, cancelling), so any expansion of it assigns $A\,\vx_{L+1}$ a coefficient of zero.
The projection is instead $c_b/\sqrt{\mathrm{Var}\,\tau+2\,\mathbb E\tau^2/L}$ with $c=\Tr_1\tilde\kappa_3+2\tilde\Sigma\tilde\mu$ (\cref{apptable:eigcoeffs}), which is nonzero whenever $c\neq0$, since $A$ correlates with functions of $W$ and $S$.
For a centered Gaussian prior $c=0$, so this mode is specific to the memorizing solution.

\subsection{Generalizing vs memorizing solution}
\label{appsubsec:dmmsevsridge}

The dMMSE is a functional of the pool measure, and each of its eigencoefficients depends on finitely many pool moments, which are polynomials in the cumulants.
Setting $\hat\kappa_{\ell\ge3}=0$ throughout is therefore the same as replacing the pool by the Gaussian with the same mean and covariance.

\begin{proposition}[The generalizing solution is a cumulant truncation of the memorizing one]
    Replacing the pool $\mathcal B_{\mathrm{tr}}$ in \cref{eq:dmmse-weights} by $\mathcal N(\hat\mu_\vbeta,\hat\Sigma_\vbeta)$ gives
    \begin{equation}
        \left(W + \epsilon^2\hat\Sigma_\vbeta^{-1}\right)^{-1}\left(S+\epsilon^2\hat\Sigma_\vbeta^{-1}\hat\mu_\vbeta\right),
    \end{equation}
    which is the ridge solution \cref{eq:ridge} with the prior's mean and covariance replaced by the pool's.
    When $\hat\Sigma_\vbeta$ is singular ($T\le d$), $\hat\Sigma_\vbeta^{-1}$ is infinite off its range, and the predicted task is pinned to $\hat\mu_\vbeta$ in directions the pool does not span.
    \label{prop:cumulant-truncation}
\end{proposition}

\begin{proof}
    Under a Gaussian prior and the linear-Gaussian likelihood, the posterior over $\vbeta$ is Gaussian, and its mean is the displayed expression.
\end{proof}

The two oracles therefore differ in exactly two ways.
The higher cumulants $\hat\kappa_{\ell\ge3}$ appear only in the dMMSE.
The first two cumulants appear in both, as $(\hat\mu_\vbeta,\hat\Sigma_\vbeta)$ versus $(\vmu_\vbeta,\Sigma_\vbeta)$. Their difference shrinks as $T^{-1/2}$ entrywise, but is severe for $T\le d$, where the pool does not span task space.

\end{document}